\documentclass[5p]{elsarticle}
\usepackage[T1]{fontenc}
\usepackage{graphicx}
\usepackage{array}   
\graphicspath{{figs/}}
\usepackage{amsmath,amssymb}
\usepackage{booktabs}
\usepackage{multirow}
\usepackage{xcolor}
\usepackage{subcaption}
\usepackage{enumitem}
\usepackage[hidelinks]{hyperref}
\usepackage{cleveref}
\usepackage{placeins}   
\usepackage{dblfloatfix} 
\usepackage{flafter}    

\biboptions{sort&compress}
\journal{Neurocomputing}

\newcommand{\R}{\mathbb{R}}
\newcommand{\E}{\mathbb{E}}
\newcommand{\norm}[1]{\left\| #1 \right\|}
\newcommand{\zA}{\mathbf{z}_A}

\newcommand{\zt}{\mathbf{z}_t}
\newcommand{\veps}{\boldsymbol{\epsilon}}
\newcommand{\vF}{v_{\mathrm{F}}}
\newcommand{\vC}{v_{\mathrm{C}}}
\newcommand{\Lcal}{\mathcal{L}}

\newcommand{\diagc}{\mathbf{c}_{\mathrm{diag}}}

\newcommand{\Tau}{\mathcal{T}}
\newcommand{\taumin}{\tau_{\min}}
\newcommand{\taumax}{\tau_{\max}}
\newcommand{\method}{PRISM}
\newcommand{\Mtgt}{\mathcal{M}_B}
\newcommand{\gridfig}[4]{%
  {\setlength{\tabcolsep}{0pt}\renewcommand{\arraystretch}{1.05}%
   \begin{tabular}{@{}*{#2}{>{\centering\arraybackslash}p{\dimexpr#1/#2\relax}}@{}}
   #3\\
   \end{tabular}}\\[1pt]%
  \includegraphics[width=#1]{#4}%
}

\usepackage{lmodern}
\usepackage{microtype}
\begin{document}

\begin{frontmatter}

\title{PRISM: Distribution-Gated Flow Matching for Controllable Unpaired Image Translation}

\author[tau]{Elad Yoshai}
\cortext[cor1]{Corresponding author.}
\author[tau]{Natan T. Shaked\corref{cor1}}
\ead{nshaked@tau.ac.il}

\affiliation[tau]{organization={School of Biomedical Engineering, Faculty of Engineering, Tel Aviv University},
                  city={Tel Aviv}, postcode={6997801}, country={Israel}}

\begin{abstract}
Unpaired image-to-image translation must decide, per image, what to change and what to preserve
without paired supervision. Many diffusion-based unpaired translators control preservation through a
single global noise or guidance value applied across the image, which cannot separate content to keep
from appearance to change. We present \method{}, a GAN-free flow-matching framework that replaces this global control with a
learned per-feature gate. The gate's spatial prior is derived from each source feature's standardized
distance to the target feature distribution, so features far from the target are freed while
target-consistent features are preserved. The same gate controls both the
initialization, which mixes the real source latent with a task-matched corruption, and the transport
timing during Ordinary Differential Equation (ODE) integration. The corruption is matched to the task, content-anchored (AdaIN) for
structure-preserving translation and partially anchored for structure-changing translation, and the gate
can be overridden locally at inference from text or a detector without retraining, preserving important structures of the original image while still generating realistic results. We evaluate \method{}
on five natural and biomedical benchmarks (AFHQ cat$\to$dog, CelebA-HQ appearance translation,
day$\to$night relighting, virtual staining, and breast frozen$\to$permanent histopathology). Among the
evaluated methods under a shared same-split protocol, \method{} attains the best Inception FID and KID on four
benchmarks and a competitive result on the fifth, and on histopathology yields the nuclei-count ratio
closest to the ideal, supporting a favorable balance between target realism and structural preservation.
\end{abstract}



\end{frontmatter}

\section{Introduction}
\label{sec:intro}
Image-to-image translation maps an image from a source domain to a target domain while
preserving its semantic content, with applications from style transfer to medical image
harmonization~\cite{isola2017image,zhu2017unpaired}. The unpaired setting, in which no
pixel-aligned source-target pairs are available, is fundamentally under-constrained. Many target images
could plausibly correspond to a given source. The central, recurring difficulty is therefore
not realism alone but control: deciding, per image and per region, which features should
change and which must be kept.

The dominant paradigm, adversarial training with cycle consistency (CycleGAN~\cite{zhu2017unpaired})
or contrastive objectives (CUT~\cite{park2020contrastive}), conflates this decision inside a
discriminator and suffers from mode collapse and training instability~\cite{goodfellow2014generative}.
Diffusion and flow editors offer a more stable alternative. SDEdit~\cite{meng2022sdedit} translates
by partially noising the source and denoising toward the target. But SDEdit exposes a single
global knob, the noise level, which governs both how much structure is destroyed and how much
style may change. Too little noise leaves the output source-like, too much destroys structure, and
the trade-off is applied uniformly to every pixel. Real translations are heterogeneous. Staining
must change while nuclear outlines must not, so the optimal control is a field rather than a scalar.

We recast unpaired translation as selective per-feature transport to the target distribution.
Each feature should change in proportion to how far its own feature statistics sit from the target
distribution, neither too much (which hallucinates) nor too little (which under-translates). We replace
the single global noise level with a distribution-informed channel-spatial gate $\tau$. High $\tau$
preserves a feature, low $\tau$ frees it. A spatial prior derived from each feature's standardized
distance to the target distribution guides the learned gate, which is then applied consistently to both
the trajectory initialization and the feature-wise timing of transport. Built over
a frozen flow-matching backbone with a norm-constrained residual correction, this yields per-feature,
distribution-informed, GAN-free translation.

\paragraph{Our contributions} We make three contributions, a central distribution-informed per-feature
preservation gate and two mechanisms that put it to work:
\begin{enumerate}[leftmargin=1.4em]
\item \textbf{A distribution-informed per-feature preservation gate (DD$\tau$)} (\Cref{sec:ddtau}).
Rather than assume or hand-tune how much each feature may change, a spatial prior is set from the
feature's standardized distance to the target distribution (with a per-image robust normalization) and
supplied as a soft target for a learned channel-spatial gate under joint training. Because the gate is
overridable at inference from a text prompt or a domain detector without retraining, one trained model
exposes controllable preservation (\Cref{sec:controllability}).
\item \textbf{The same gate controls both initialization and transport timing}
(\Cref{sec:taupredictor}). High-$\tau$ features are initialized from the real source latent and receive
near-zero transport, so preserved content stays at its source value, and the same gate governs when each
feature is allowed to move during integration.
\item \textbf{Task-matched initialization} (\Cref{sec:taupredictor}). A single design switch at $t{=}0$,
$\mathrm{AdaIN}$(source content, target style) for structure-preserving tasks and a partially-anchored
isotropic corruption for structure-changing ones, adapts the framework to both regimes.
\end{enumerate}
Across five benchmarks spanning
both regimes, three structure-preserving (breast frozen$\to$permanent, day$\to$night, virtual staining)
and two structure-changing (AFHQ cat$\to$dog, face man$\to$woman), \method{} attains the best Inception FID and KID
among the compared methods on four and a competitive result on the fifth, and on histopathology yields
the nuclei-count ratio closest to ideal. A realism-faithfulness analysis (\Cref{sec:frontier}) shows this
reflects selective transport rather than under-translation (\Cref{sec:medical}).

\section{Related Work}
\label{sec:related}
\paragraph{Unpaired translation} CycleGAN~\cite{zhu2017unpaired} made training without paired
examples possible through a cycle-consistency constraint, where translating an image to the other
domain and back should recover the original. This ties its two one-directional generators together and
rules out arbitrary mappings that ignore the input. MUNIT~\cite{huang2018multimodal} extended the idea
to multimodal outputs by factoring each image into a domain-shared content code and a domain-specific
style code that can be resampled to yield several outputs per input, and StarGAN-v2~\cite{choi2020stargan} translates among many domains with a
single generator driven by learned per-domain style vectors. CUT~\cite{park2020contrastive} removed the
cycle and its second generator, replacing them with a patch-wise contrastive objective that maximizes
mutual information between corresponding input and output patches, so each output patch is pulled toward
its own source location and away from the rest. Attention-guided and masked-discriminator methods
further localize translation and reduce unintended content changes~\cite{tang2023attentiongan,stuhr2026masked}.
Pix2pix~\cite{isola2017image} addresses the paired
setting with a conditional GAN and therefore requires pixel-aligned image pairs, so it is not directly comparable to the unpaired methods studied here. These adversarial methods are
susceptible to mode collapse and training instability~\cite{goodfellow2014generative}.
\method{} avoids adversarial training entirely.

\paragraph{Diffusion and flow translation} Diffusion models~\cite{ho2020denoising,song2021scorebased}
and their deterministic samplers~\cite{song2021denoising} underpin SDEdit~\cite{meng2022sdedit}, latent
diffusion~\cite{rombach2022high}, and ControlNet~\cite{zhang2023adding} (for paired images). SDEdit translates by
adding a controlled amount of noise to the source and then denoising it under the target-domain model,
so a single global noise level sets the balance between preserving structure and changing appearance.
Leading unpaired diffusion translators, EGSDE~\cite{zhao2022egsde}, SDDM~\cite{sun2023sddm} and
UNSB~\cite{kim2023unpaired}, steer a target-domain process toward realism while retaining source
content. EGSDE guides the reverse-time SDE with two energy terms,
a domain-independent expert that keeps the output close to the source structure and a domain-specific
expert that pushes it onto the target domain, combined through a scalar weight. SDDM
decomposes the score on the data manifold to separate style from content and edits only the style
component. Schr\"odinger-bridge formulations~\cite{shi2024diffusion}, and their unpaired extension
UNSB, learn a stochastic bridge that carries one domain's distribution to the
other in a few stochastic steps, casting unpaired translation as entropic optimal transport. We
benchmark SDEdit, EGSDE, and UNSB as representative instances of this family (SDDM is discussed for
context but not re-implemented). These diffusion translators share three structural limitations that
\method{} addresses. First, preservation is governed by a \textbf{single global} quantity, a noise level
(SDEdit), a guidance weight (EGSDE), or a decomposition coefficient (SDDM), applied uniformly across
the image, so they cannot decide preservation per feature. Second, their inference-time controls are
global, whereas \method{} uses a learned per-feature gate that can be overridden locally without
retraining. Third, because they
renoise the source under a \textbf{single global noise level} rather than a per-feature gate, content
that should be preserved is still partially resampled and can drift or be hallucinated. \method{}
initializes the preserved features from the real source and applies near-zero transport there, so those
features remain strongly source-anchored rather than being resampled. CycleGAN-Turbo~\cite{parmar2024cyclganturbo}
attaches lightweight adapters to a pretrained text-to-image diffusion model for one-step unpaired
translation. \method{} shares the frozen-backbone-plus-learned-correction structure, but its learned
component is a per-feature preservation gate that decides what to keep rather than a general
domain adapter, and it needs no pretrained text-to-image model. Flow matching~\cite{lipman2023flow}, rectified
flow~\cite{liu2023flow} and optimal-transport (OT)-conditional flow matching~\cite{tong2024improving} provide our
straight-path backbone. \Cref{tab:positioning} compares \method{} with these baselines by mechanism. The individual ingredients here are established, namely an SDEdit-style initialization, an AdaIN
corruption, and learned masks. What is specific to \method{} is coupling a distribution-derived
preservation field to both the initialization and the transport timing inside a single trained gate, the
property the table isolates.

\begin{table*}[t]
\centering\small
\caption{Positioning of \method{} against the compared unpaired translators, by mechanism. Cells state
the kind of control rather than a binary mark. ``Preservation control'' is Global (one value for the
whole image), Implicit (learned inside a generator and not exposed), or Per-feature. ``Gate on
initialization'' and ``Gate on transport timing'' indicate whether a per-feature gate sets the ODE start
point and the timing of transport. ``Local inference-time control'' is whether preservation can be
adjusted locally at inference without retraining.}
\label{tab:positioning}
\setlength{\tabcolsep}{6pt}
\begin{tabular}{lcccccc}
\toprule
Property & CycleGAN & CUT & SDEdit & EGSDE & UNSB & \method{} (Ours)\\
\midrule
GAN-free training & No & No & Yes & Yes & Yes & Yes\\
Preservation control & Implicit & Implicit & Global & Global & Global & Per-feature\\
Prior from distribution distance & No & No & No & No & No & Yes\\
Gate on initialization & No & No & Global & No & No & Per-feature\\
Gate on transport timing & No & No & No & No & No & Yes\\
Source-initialized preserved content & No & No & No & No & No & Yes\\
Local inference-time control & No & No & Global & Global & No & Learned\\
\bottomrule
\end{tabular}
\end{table*}

\paragraph{Controllability and preservation} A recurring difficulty is controlling what is
preserved versus changed. Spatially varying control typically needs user masks or attention edits, and
none learn the preservation map from the data distributions themselves. Closer to spatial control, ILVR~\cite{choi2021ilvr} guides a diffusion
process toward a reference at a hand-set low-pass scale, blended diffusion~\cite{avrahami2022blended}
and DiffuseIT~\cite{kwon2023diffuseit} edit inside a user-provided mask or under a global
guidance term, and bridge formulations such as DDIB~\cite{su2023ddib} and I$^2$SB~\cite{liu2023i2sb}
connect the two domains with stochastic bridge that applies the same
transport everywhere. Closest to our data-driven map, DiffEdit~\cite{couairon2023diffedit}
automatically infers an edit mask by contrasting prompt-conditioned and unconditioned diffusion
noise estimates, then inpaints inside it. It differs from \method{} in three ways. Its mask is binary
and text-prompt-derived rather than a continuous gate measured from each feature's distributional
distance to the target, it drives masked inpainting rather than being integrated into the transport
dynamics, and it targets a paired text-to-image editor rather than unpaired distribution translation.
Prior spatially-varying editors
and edit-mask methods take the preservation region as given, from a user mask, a text prompt, or a
hand-set scale, and mark only \emph{where} to edit. \method{} instead \emph{measures} preservation. The
field is read from each feature's standardized distance to the target distribution and sets \emph{how
much} each feature moves, and the same measured field is applied at two coupled points, the
corruption-mixed initialization and the wake-up timing of transport. \method{} thus differs from recent
spatially adaptive diffusion and flow approaches by deriving its preservation prior from
source-to-target feature-distribution discrepancy and applying the resulting gate jointly to
initialization and transport timing.

\paragraph{Histopathological translation} Frozen-section artifacts challenge intraoperative diagnosis~\cite{tarek2023frozen}. Virtual staining and related
information-preserving biomedical translation methods~\cite{dehaan2021deep,rivenson2019virtual,you2025preserving},
as well as style-transfer approaches~\cite{ozyoruk2022deep,levy2024staintransfer}, help but often need
paired images or are limited to specific tissue types. Prior work restores frozen sections with permanent-section-guided learning and
nuclei attention~\cite{yoshai2025enhancing}. Pathology foundation models such as Phikon~\cite{filiot2023scaling,filiot2024phikon},
UNI~\cite{chen2024towards}, and CONCH~\cite{lu2024conch}, provide strong feature extractors we use for
evaluation. Our diagnosis-conditioned guard exploits The Cancer Genome Atlas (TCGA)~\cite{tcga2013cancer} clinical metadata to
anchor semantic content during unpaired translation.

\section{Method}
\label{sec:method}
\begin{figure*}[t]
\centering
\includegraphics[width=0.92\linewidth]{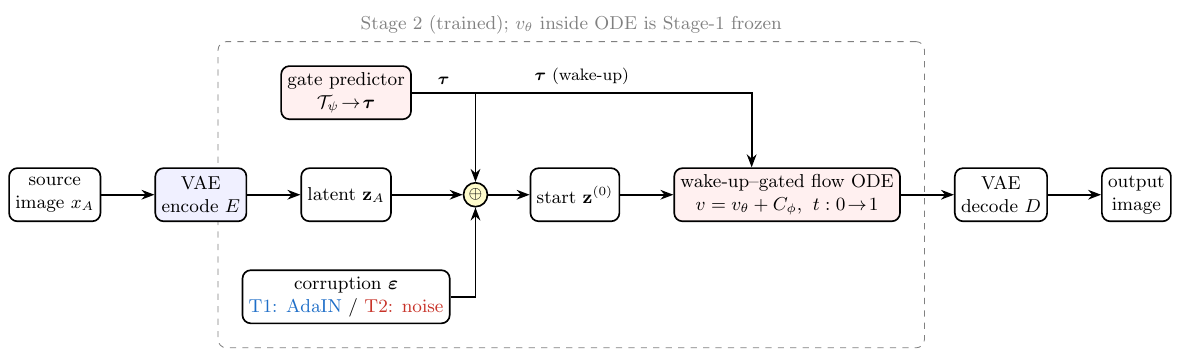}
\caption{\textbf{Overview of \method{}'s distribution-informed gated transport.} The predictor
$\mathcal{T}_\psi$ produces a channel-spatial gate $\tau$ from the source latent $\mathbf{z}_A$, guided
by a spatial prior derived from source-to-target feature-distribution discrepancy (DD$\tau$,
\Cref{sec:ddtau}). The \emph{same} gate $\tau$ enters the pipeline twice. It mixes the real source
latent with the task-matched corruption $\veps$ (content-anchored AdaIN for structure-preserving tasks,
partially-anchored noise for structure-changing ones) to form the start point $\mathbf{z}^{(0)}$
(\Cref{eq:freq_init}), and it sets the feature-wise wake-up timing of transport in the ODE
(\Cref{eq:gated_vel}). High-$\tau$ features stay source-anchored and transport little, while low-$\tau$
features start closer to the corruption and transport earlier. Purple = trained in Stage~1 then frozen
($v_\theta$). red = trained in Stage~2 ($\mathcal{T}_\psi$ and the norm-constrained correction $C_\phi$).
The result is decoded to the final output.}
\label{fig:pipeline}
\end{figure*}
\method{} is a two-stage framework for unpaired translation as selective per-feature transport to the
target distribution, controlled by a learned, distribution-driven gate field $\tau$. Let
$\mathcal{X}_A,\mathcal{X}_B$ be source and target domains with distributions $p_A,p_B$. We operate in
the latent space of a frozen Stable-Diffusion variational autoencoder (VAE)~\cite{rombach2022high}, $\mathbf{z}=\mathcal{E}(x)\in\R^{C\times H\times W}$
with $C{=}4$, $H{=}W{=}32$ for the $256{\times}256$ inputs used throughout (the VAE downsamples
$8\times$). \Cref{fig:pipeline} shows how the frozen flow, the gate predictor,
and the correction network compose to translate one image at inference. \Cref{tab:components} lists every
component, its role, whether it is trained, frozen, or evaluation-only, the tasks it applies to, and
whether it is central to the mechanism or a supporting term.

\begin{table*}[t]
\centering\small
\caption{Components of \method{}: each component's role, training state, and task scope. State is
trained (in Stage~2), frozen (pretrained then held fixed), precomputed (once per dataset), or a fixed
rule. The gated-transport core ($v_\theta$, $\veps_\alpha$, $\Tau_\psi$, DD$\tau$ prior, $C_\phi$) is
complemented by supporting realism and consistency terms (MMD, DINO matcher, diagnosis guard, auxiliary
preservation) and an optional learned initialization (LNG).}
\label{tab:components}
\setlength{\tabcolsep}{8pt}
\begin{tabular}{llll}
\toprule
Component & Role & State & Tasks \\
\midrule
Stage-1 flow $v_\theta$ & domain-conditional generative flow & frozen & all \\
Corruption $\veps_\alpha$ & task-matched $t{=}0$ initialization & fixed rule & all \\
Gate predictor $\Tau_\psi$ & channel-spatial preservation gate $\tau$ & trained & all \\
DD$\tau$ prior & distribution-derived spatial prior for $\tau$ & precomputed & all \\
Correction $C_\phi$ & norm-bounded on-support correction & trained & all \\
Inception MMD & GAN-free realism driver & trained & all \\
$k$-NN DINO matcher & per-region realism & frozen encoder & all \\
Diagnosis guard $h_\psi$ & semantic consistency & frozen & medical \\
Aux.\ preservation terms & Type-1 stabilization (supporting) & trained & Type-1 \\
LNG $G_\xi$ (optional) & deterministic initialization & trained & optional \\
\bottomrule
\end{tabular}
\end{table*}

\subsection{Stage 1: Domain-Conditional Flow Matching (Frozen)}
\label{sec:stage1}
Stage~1 learns, once, a generative flow that maps Gaussian noise to each domain, which Stage~2 then
reuses frozen. We adopt the linear-interpolant flow matching of~\cite{lipman2023flow}. For a clean
latent $\mathbf z$, the encoding of a real image, and Gaussian noise $\veps\sim\mathcal N(0,I)$, we form
the straight-line interpolant $\zt=(1-t)\veps+t\,\mathbf z$ at a random time $t\sim\mathcal U(0,1)$,
whose constant velocity toward the data is $v^\star=\mathbf z-\veps$. A domain-conditional network
$v_\theta(\zt,t,d)$ is trained to predict this velocity, where $d$ is the domain label. The training
objective is the flow-matching loss
\begin{equation}
\label{eq:fm_loss}
\Lcal_{\mathrm{FM}}=\E_{d,\;\mathbf z\sim p_d,\;\veps,\;t}\bigl[\norm{v_\theta(\zt,t,d)-(\mathbf z-\veps)}^2\bigr],
\end{equation}
where $\E$ is the expectation
over domains $d$, real latents $\mathbf z\sim p_d$, noise $\veps$,
and time $t$, and $\mathbf z-\veps$ is the regression target.
After training we freeze $v_\theta$. Integrating it from $t{=}0$ to $t{=}1$ with the target label $d{=}B$
carries a Gaussian sample onto the target domain, and we write $\Mtgt$ for the set of latents it produces, the target set. Freezing the flow helps Stage~2 in two ways: $\Mtgt$ is now
fixed, so Stage~2 optimizes against a stationary target rather than a moving one, and Stage~2 no
longer has to learn the target distribution, only to correct the off-manifold drift caused by
starting the ODE from a corrupted source instead of from pure noise.

\subsection{Per-Feature Transport: the $\tau$ Gate}
\label{sec:ot}
\method{} uses a per-feature gate field
$\tau\in[\taumin,\taumax]^{C\times H\times W}$ ($\taumin{\approx}0.05$, $\taumax{=}1$), one scalar per
latent feature-location and therefore the same shape as the latent. At each location, it quantifies how
strongly the feature should be preserved ($\tau{\to}1$) versus translated ($\tau{\to}0$).

The gate's preservation prior derives from an idealized transport target. Given the frozen target set $\Mtgt$, the
ideal output is the point on $\Mtgt$ reachable from the source $x_A$ by changing each feature as little
as its own distribution requires:
\begin{equation}
\label{eq:ot_obj}
T^\star(x_A)=\arg\min_{\hat x:\,\mathcal{E}(\hat x)\in\Mtgt}\;\sum_p \tau(p)\,\big\|\mathcal{E}(\hat x)(p)-\mathcal{E}(x_A)(p)\big\|^2 .
\end{equation}
Here $\hat x$ is a candidate output, $\mathcal E(\cdot)$ the VAE encoder, $p$ indexes feature-locations,
and $\mathcal E(\hat x)(p)$ and $\mathcal E(x_A)(p)$ are the output and source features at location $p$.
The weight $\tau(p)$ is the per-feature cost of moving that feature. High $\tau$ is expensive and
preserved, low $\tau$ cheap and freed. The constraint
$\mathcal E(\hat x)\in\Mtgt$ keeps the output on the target set.

We treat \Cref{eq:ot_obj} as a design principle that names the role of each component, not as a program
we solve exactly. Because the frozen flow supplies the drive toward target realism, $\tau$
controls how far along the source-to-target path each feature is allowed to travel. Two
signals set $\tau$. The per-task corruption regime (content-anchored vs.\ isotropic, \Cref{eq:alpha_blend})
fixes the overall preserve/translate balance, and the distribution-driven prior DD$\tau$
(\Cref{sec:ddtau}) supplies its per-feature spatial pattern.

\subsection{The Gate Predictor and the Gated Sampler}
\label{sec:taupredictor}
A small ($\sim$3M-parameter) U-Net $\Tau_\psi$, the gate
predictor, reads the source latent $\zA$ and emits one gate value per feature-location,
\begin{equation}
\label{eq:predictor}
\tau=\taumin+(\taumax-\taumin)\,\sigma\!\big(\Tau_\psi(\zA)\big),
\end{equation}
where $\Tau_\psi(\zA)$ is the predictor's raw output, $\sigma$ the logistic sigmoid, and the affine
rescaling by $\taumin{\approx}0.05$ and $\taumax{=}1$ puts $\tau\in[\taumin,\taumax]$ by construction.
We zero-initialize $\Tau_\psi$, so training starts
from a neutral gate $\tau_{\text{init}}$ everywhere and the losses then shape it.

The gate enters the sampler in two places. The first is the initialization. Instead of starting the
Stage-2 ODE from pure noise, we start from a per-feature mixture of the real source and a corruption
$\veps_\alpha$, weighted by the gate:
\begin{equation}
\label{eq:freq_init}
\mathbf z^{(0)}=\tau\odot\zA+(1-\tau)\odot\veps_\alpha ,
\end{equation}
where $\mathbf z^{(0)}$ is the ODE start point, $\odot$ denotes elementwise multiplication, $\zA$ is the
source latent, and $\veps_\alpha$ is the task-matched corruption defined in
\Cref{eq:content_anchored,eq:alpha_blend}. This is a per-feature generalization of
SDEdit. Where SDEdit renoises the whole image by one global amount, \Cref{eq:freq_init} renoises each
feature by its own amount $1-\tau$. High-$\tau$ features start from the real source latent and receive
near-zero transport, so they are preserved, while low-$\tau$ features start mostly from the corruption and are
painted into target content. In the $\tau{\to}1$ limit a feature is initialized at the source latent,
keeping preserved regions strongly source-anchored rather than resampling them.

\begin{figure}[t]
\centering
\gridfig{0.99\linewidth}{5}{\small source & \small IN & \small target style & \small content-anchored & \small isotropic}{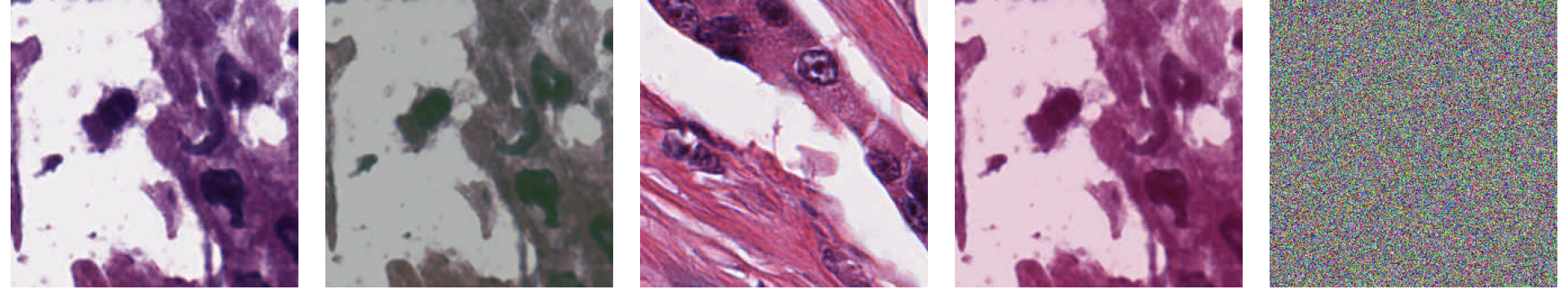}
\caption{\textbf{Task-matched $t{=}0$ corruption} (shown in pixel space for clarity, computed on the
latent). Columns, left to right. The source $\mathbf{z}_A$. Instance normalization
$\mathrm{IN}(\mathbf{z}_A)$, which retains spatial structure and drops global colour. A target sample
supplying the per-channel style $(\tilde\mu,\tilde\sigma)$. The content-anchored corruption
$\veps_{\text{struct}}$, source structure with target style, whose
$\tau{\to}0$ fixed point is a valid style transfer, and the isotropic noise $\veps\sim\mathcal N(0,I)$.}
\label{fig:corruption}
\end{figure}

The task-matched corruption $\veps_\alpha$ in \Cref{eq:freq_init} is not always isotropic.
Isotropic noise is
correct when the freed structure must be re-synthesized, e.g.\ cat$\to$dog, where the
geometry itself (snout, ears, face shape) changes. For structure-preserving translation,
such as virtual staining, frozen$\to$permanent histology, day$\to$night, the spatial layout is a
shared invariant. Only appearance (color, stain, texture) should move. There, isotropic $\veps$
needlessly destroys content in exactly the regions $\tau$ frees, forcing the flow to hallucinate the
geometry back. We instead corrupt toward the target style while
anchoring the source content. Let $\mu_c,\sigma_c$ be the per-channel mean/std of $\zA$ over spatial
positions and $\mathrm{IN}(\zA)=(\zA-\mu_c)/\sigma_c$ the instance-normalized latent, which discards
global appearance but retains the spatial pattern (the structure). Drawing a per-channel target style
$(\tilde\mu,\tilde\sigma)$ from a target-domain bank, we replace $\veps$ in \eqref{eq:freq_init} with
\begin{equation}
\label{eq:content_anchored}
\veps_{\text{struct}}=\mathrm{IN}(\zA)\odot\tilde\sigma+\tilde\mu,\qquad
\tilde\sigma\leftarrow\max(\tilde\sigma,\,c\,\sigma_c).
\end{equation}
At $\tau\!\to\!0$ this is exactly Adaptive Instance Normalization (AdaIN)~\cite{huang2017arbitrary}. The
corruption's fixed point is a valid style transfer rather than noise, so structure is preserved at
every $\tau$ and the DD$\tau$ gate (\Cref{sec:ddtau}) transports only appearance, reducing the usual trade-off between preserving structure and changing style that arises when both are controlled by a single global noise level. The clamp
$\tilde\sigma\!\ge\! c\,\sigma_c$ ($c{=}0.85$) is a one-sided contrast floor. A low-contrast style draw
cannot desaturate the output below a fraction of the source's own dynamic range, removing washed-out
failures on pale regions while leaving well-saturated draws untouched. AdaIN itself is standard. Our
contribution is its use as the $t{=}0$ corruption \emph{inside} flow transport, so one trained model
preserves structure while the gate moves only appearance (\Cref{fig:corruption}).

The isotropic corruption and the content-anchored one
are the two endpoints of a one-parameter family, which we blend at $t{=}0$,
\begin{equation}
\label{eq:alpha_blend}
\veps_\alpha=\alpha\,\veps_{\text{struct}}+(1-\alpha)\,\veps,\qquad \alpha\in[0,1],
\end{equation}
where $\veps_\alpha$ is the blended corruption actually used in \Cref{eq:freq_init},
$\veps_{\text{struct}}$ is the content-anchored corruption, $\veps$ is the isotropic one, and the mixing
weight $\alpha$ is the corruption's anchor strength. Because $\mathrm{AdaIN}$ matches the source's
spatial pattern to the target's per-channel mean and variance, $\veps_{\text{struct}}$ approximates the
optimal-transport map between the source and target feature distributions under a per-channel
diagonal-Gaussian model. This first-order map ignores cross-channel and higher-order structure, and the
Stage-2 correction $C_\phi$ and the DD$\tau$ gate close the remaining gap, so we need no exact joint-OT
solve. $\alpha$ interpolates the corruption toward this fixed point. Frozen$\to$permanent and day$\to$night translation use $\alpha{=}1$ to preserve geometry, while virtual
staining uses the lighter anchor $\alpha{=}0.3$ described in \Cref{sec:experiments}. For structure-changing tasks the binary choice is
suboptimal. Pure isotropic noise ($\alpha{=}0$) discards the source, so the synthesized dog need not
resemble the input cat, whereas a full anchor would freeze the cat geometry. A partial anchor
($\alpha{=}0.5$) carries the source's coarse appearance and pose into the start point while
leaving low-$\tau$ features available for the flow to re-synthesize structure, a real dog that still
resembles the input cat. $\alpha$ thus exposes a controllable trade-off between distribution fidelity and
input faithfulness. On AFHQ, raising $\alpha$ from $0$ to $0.5$ increases the source-cat-to-output-dog
CLIP (Contrastive Language-Image Pre-training) cosine similarity from $0.58$ to $0.65$, at only a modest cost in FID, a frontier
we report rather than a single operating point (\Cref{sec:afhq}). Closeness to the input is supplied by the corruption itself, and the
DD$\tau$ gate (\Cref{sec:ddtau}) preserves it (\Cref{fig:corruption}).

The second place the gate enters is the integration. We integrate the ODE with a composite velocity
$\mathbf v=\vF+\vC$, where $\vF=v_\theta(\mathbf z^{(k)},t_k,B)$ is the frozen base velocity at step $k$
and $\vC$ is the learned correction (\Cref{sec:correction}). A per-feature wake-up gate then
decides how early each feature is allowed to move:
\begin{equation}
\label{eq:gated_vel}
\mathbf v_{\text{gated}}=\sigma\!\big((t_k-\tau)/T\big)\odot\mathbf v,\qquad
\mathbf z^{(k+1)}=\mathbf z^{(k)}+\Delta t\cdot\mathbf v_{\text{gated}},
\end{equation}
where $\mathbf v_{\text{gated}}$ is the gated velocity, $\mathbf z^{(k)}$ the state at step $k$, $t_k$
the current time, $\Delta t$ the step size, and $\sigma\!\big((t_k-\tau)/T\big)$ a soft per-feature
switch that turns on only once $t_k$ passes the feature's own gate value $\tau$, with sharpness
$T{=}0.15$. We take $K$ Euler steps ($K{=}4$ in training, $K{=}16$ at inference, with quality broadly
stable across moderate $K$, though additional steps do not consistently improve FID). A high-$\tau$ feature remains inactive through most of the trajectory,
while a low-$\tau$ feature becomes active earlier and translates sooner.

Both the initialization (\Cref{eq:freq_init}) and the wake-up integration (\Cref{eq:gated_vel}) are
smooth functions of $\tau$, because the wake-up is a soft sigmoid gate rather than a hard threshold.
This smoothness makes the per-feature gate trainable end to end. Gradients from the realism
objectives flow back through the decoded output and the ODE steps into the predictor $\Tau_\psi$. The
gate is therefore not merely regressed onto the DD$\tau$ target but also shaped by the end task,
learning to open (low $\tau$) only where translating actually improves target realism and to stay
closed (high $\tau$) elsewhere. A hard threshold would zero the gradient with respect to $\tau$ and
break this, whereas the sharpness $T$ trades gradient flow against gate crispness, with $T\!\to\!0$
recovering a hard SDEdit-style cutover. The three roles are thus separated yet jointly
optimized. The frozen flow supplies the base trajectory, the correction $C_\phi$ the deflecting
velocity, and $\Tau_\psi$ learns per feature where and when it is applied. \Cref{fig:sampling} gives
the full inference procedure in pseudo-code.

\begin{figure}[t]
\centering
\setlength{\fboxsep}{6pt}
\fbox{\begin{minipage}{0.9\linewidth}\small
\textbf{PRISM sampling}~($x_A\to$ target $B$)\\[2pt]
\textbf{Require:} source $x_A$, frozen $v_\theta,C_\phi,\Tau_\psi$, steps $K$ with $\Delta t{=}1/K$, sharpness $T$, and task regime\\[3pt]
\resizebox{\linewidth}{!}{$\begin{array}{@{}r@{~}l@{}}
1: & \zA \gets \mathcal{E}(x_A)\ \text{\footnotesize(encode)}\\[2pt]
2: & \tau \gets \taumin+(\taumax-\taumin)\,\sigma(\Tau_\psi(\zA))\ \text{\footnotesize(gate, Eq.~\ref{eq:predictor})}\\[2pt]
3: & \text{build }\veps_\alpha:\ \veps_{\text{struct}}\ \text{(Type-1) or isotropic/blend (Type-2), Eq.~\ref{eq:content_anchored},\ref{eq:alpha_blend}}\\[2pt]
4: & \mathbf z^{(0)} \gets \tau\odot\zA+(1-\tau)\odot\veps_\alpha\ \text{\footnotesize($\tau$-gated init, Eq.~\ref{eq:freq_init})}\\[2pt]
5: & \textbf{for }k=0\ \textbf{to }K-1\textbf{:}\\[2pt]
6: & \quad t_k \gets k\,\Delta t,\quad \vF \gets v_\theta(\mathbf z^{(k)},t_k,B)\\[2pt]
   & \quad \vC \gets C_\phi(\cdot)\ \text{\footnotesize(clipped correction, Eq.~\ref{eq:norm_constraint})}\\[2pt]
7: & \quad \mathbf v_{\text{gated}} \gets \sigma\!\big((t_k-\tau)/T\big)\odot(\vF+\vC)\ \text{\footnotesize(wake-up, Eq.~\ref{eq:gated_vel})}\\[2pt]
8: & \quad \mathbf z^{(k+1)} \gets \mathbf z^{(k)}+\Delta t\cdot\mathbf v_{\text{gated}}\\[2pt]
9: & \textbf{return }\hat x_B \gets \mathcal{D}(\mathbf z^{(K)})\ \text{\footnotesize(decode)}
\end{array}$}
\end{minipage}}
\caption{The \method{} inference procedure (pseudo-code). All learned modules ($v_\theta,C_\phi,\Tau_\psi$)
are frozen during sampling, and only the gate $\tau$ can be overridden at inference.}
\label{fig:sampling}
\end{figure}

\subsection{$\tau$-Adaptive Per-Feature Compute}
\label{sec:adaptive}
The wake-up gate enables a built-in efficiency opportunity. A feature with gate $\tau(p)$ contributes
essentially no velocity until $t_k\gtrsim\tau(p)$. Fixing a single step size $\Delta t$, we integrate
each feature only over its active interval $[\tau(p),1]$, giving a per-feature number of function
evaluations (NFE)
\begin{equation}
\label{eq:adaptive_nfe}
\mathrm{NFE}(p)=\big\lceil(1-\tau(p))/\Delta t\big\rceil,
\end{equation}
so preserved (high-$\tau$) features cost few function evaluations and translated features receive the
full budget. The same field that controls what changes also dictates how much compute each region
receives. This localization is specific to per-feature gating, whereas single-knob editors apply
a global step count everywhere. These are idealized per-region evaluation counts, not wall-clock savings. The dense backbone still evaluates every spatial position at each step, so realizing them requires structured sparse execution, which we leave to future work. We report the per-region counts in \Cref{sec:adaptive_results}.

\subsection{Constrained Residual Correction}
\label{sec:correction}
Starting the ODE from the source-conditioned corruption (\Cref{eq:freq_init}) differs from the noise
distribution on which the frozen flow was trained. A
$\sim$17M-param Diffusion Transformer (DiT)~\cite{peebles2023scalable} $C_\phi$ (hidden 384, depth 6, patch
4) outputs an additive velocity
$\vC=C_\phi(\mathbf z^{(k)},t_k,\phi_{\mathrm{feat}}(x_A),\mathrm{dir},\diagc)$, zero-initialized and
norm-constrained, which adapts the effective vector field while limiting its magnitude relative to the
frozen velocity:
\begin{equation}
\label{eq:norm_constraint}
\vC\leftarrow\vC\cdot\min\!\Big(1,\ \frac{\beta\,\norm{\vF}}{\norm{\vC}+\veps_{\mathrm{num}}}\Big),\quad\beta=0.5.
\end{equation}
Here $\vC$ is the learned correction velocity, $\vF$ the frozen base velocity, and
$\veps_{\mathrm{num}}>0$ a small constant for numerical stability. Only the correction's
magnitude is clipped, to at most a fraction $\beta{=}0.5$ of
the base-velocity magnitude, a norm clip $\norm{\vC}\!\le\!\beta\norm{\vF}$ applied first using the
channelwise norm at each spatial location and then using the norm over the full tensor. Because $\vC$
retains its own orientation, it can adjust the frozen vector field to counter off-manifold drift while
its magnitude stays bounded relative to the base velocity, without overwriting the base flow, which
supports stability without an adversary. Since only magnitude is bounded, training is not sensitive to the exact
value of $\beta$. Any value within a broad range simply determines how aggressively the correction may deflect, and we use a fixed value of $\beta{=}0.5$.

\subsection{$k$-NN Local DINO Feature Matcher}
\label{sec:matcher}
For non-saturating, per-region realism we cache, for each source, the statistics of its $k$ nearest
real targets in frozen DINOv2~\cite{oquab2024dinov2} space and minimize a per-sample Mahalanobis
distance. Before training we extract class-token (CLS) and patch features for all sources and targets. For each source
$x_A$ we form $\mathcal N_k(x_A)$ by top-$k$ CLS cosine similarity and cache its local
mean/std $(\mu^{(i)},\sigma^{(i)})$. The loss for a translated $\hat x_B^{(i)}$ is
\begin{equation}
\label{eq:matcher_loss}
\Lcal_{\mathrm{match}}^{\mathrm{local}}=\Big\|\tfrac{\hat f_{\mathrm{cls}}-\mu^{(i)}_{\mathrm{cls}}}{\sigma^{(i)}_{\mathrm{cls}}}\Big\|_2^2+\Big\|\tfrac{\hat f_{\mathrm{patch}}-\mu^{(i)}_{\mathrm{patch}}}{\sigma^{(i)}_{\mathrm{patch}}}\Big\|_2^2.
\end{equation}
Each source has its own target statistics (superscript $(i)$). The loss translates this
specific cat toward the target region that looks most like it, with the per-sample minimal-displacement
target of \Cref{eq:ot_obj}.

\subsection{Structure Preservation and Realism}
\label{sec:cycle}\label{sec:struct}\label{sec:mmd}
Where $\tau$ is high, the output must stay faithful to the source. We impose a $\tau$-gated structure
loss,
\begin{equation}
\label{eq:struct}
\Lcal_{\mathrm{struct}}=\sum_{g\in\{\mathrm{sem},\mathrm{pix},\mathrm{edge}\}}\lambda_g\,\mathrm{mean}\big(\mathrm{sg}(\tau)\odot\delta_g(\hat x_B,x_A)\big),
\end{equation}
where $\hat x_B$ is the translated output, $x_A$ the source, $\delta_{\mathrm{sem}}$ a DINOv2 patch
distance, $\delta_{\mathrm{pix}}$ an image-space $L_1$ distance, and $\delta_{\mathrm{edge}}$ a
Sobel-gradient distance. $\mathrm{sg}(\tau)$ is the
stop-gradient gate, which weights the penalty by how much each feature should be preserved while
preventing the loss from lowering $\tau$ to escape its own penalty, and $\lambda_g$ are the
per-granularity weights.

\subsection{Distribution-Driven $\tau$ Prior (DD$\tau$)}
\label{sec:ddtau}
In \method{}, the gate's \emph{spatial} pattern is supervised by a
distribution-distance prior. With a frozen DINOv2 patch encoder $\phi$ and target moments
$(\mu_B,\sigma_B)$, for each source patch $p$,
\begin{equation}
\label{eq:ddtau_prior}
\begin{aligned}
d(p)&=\big\|(\phi(x_A)(p)-\mu_B)/\sigma_B\big\|^2,\\
\tau_{\mathrm{prior}}(p)&=1-\mathrm{norm}_q\big(d(p)\big),
\end{aligned}
\end{equation}
with per-image $q$-quantile normalization $\mathrm{norm}_q(d(p))=\min\!\big(1,\,d(p)/q\big)$, where $q$ is
the $0.95$-quantile of $\{d(p')\}_{p'}$ over the image, floored at a small positive constant for numerical
stability, so $\tau_{\mathrm{prior}}(p)\in[0,1]$. A patch far from the target distribution receives a low $\tau_{\mathrm{prior}}$ while a target-like patch receives a high $\tau_{\mathrm{prior}}$. We add a soft consistency loss
$\Lcal_{\mathrm{DD\tau}}=\norm{\tau-\mathrm{sg}(\tau_{\mathrm{prior}})}^2$, where $\tau$ is the predicted gate defined in \Cref{eq:predictor}. The
prior $\tau_{\mathrm{prior}}$ is spatial, interpolated to the latent resolution and shared across channels. It serves as a soft target for $\Tau_\psi$,
whose final output $\tau$ is channel-spatial and jointly optimized with the remaining objectives. Thus, the spatial
preservation pattern is distribution-derived, while the channel-specific refinements are learned.

This choice has a simple transport interpretation. Because $d(p)$ measures each source patch's standardized distance to the global target feature statistics $(\mu_B,\sigma_B)$, patches that are already close to the target need little transport, while distant patches require more. This is a first-order distributional proxy rather than a per-location conditional comparison. Accordingly, setting $\tau_{\mathrm{prior}}(p)$ from $d(p)$ allocates transport in proportion to each feature's estimated discrepancy from the target distribution.

\subsection{Diagnosis-Consistency Regularization for Medical Imaging}
\label{sec:diag_bridge}
A translated frozen section must preserve the source patient's tissue type and diagnosis. This is achieved through two complementary mechanisms. First, Stage~1 is conditioned on organ/diagnosis/grade metadata, where the embedding
$\diagc=[\mathbf e_{\mathrm{organ}};\mathbf e_{\mathrm{diag}};\mathbf e_{\mathrm{grade}}]$ is injected into
$v_\theta$ (and passed to $C_\phi$) via adaptive conditioning, with classifier-free dropout so it is
optional at inference. Second, Stage~2 adds a diagnostic-consistency loss using a frozen diagnosis
classifier $h_\psi$, pretrained once on permanent sections with same-patient TCGA
labels~\cite{tcga2013cancer}, then held fixed. The translated tile must receive the same predicted class
as the source tile,
\begin{equation}
\label{eq:diag}
\Lcal_{\mathrm{diag}}=\mathrm{CE}\big(h_\psi(\hat x_B),\,\arg\max h_\psi(x_A)\big),
\end{equation}
where the target label $\arg\max h_\psi(x_A)$ is detached. The Stage-2 loss itself uses no ground-truth labels, only source-vs-output agreement. The TCGA labels enter only in the one-time
pretraining of $h_\psi$.

\subsection{Total Objective and Role Decomposition}
\label{sec:objective}
Stage~1 minimizes the flow-matching loss $\Lcal_{\mathrm{FM}}$ (\Cref{eq:fm_loss}), after which $v_\theta$ is frozen.
Stage~2 then trains only the correction $C_\phi$, the feature matcher
$\phi_{\mathrm{feat}}$, and the gate predictor $\Tau_\psi$. The diagnosis classifier $h_\psi$, used for medical frozen$\to$permanent translation
only, is pretrained once and held frozen throughout Stage~2.
The objective has three parts, one per degree of freedom of the transport: a
realism term that pulls outputs onto the target set, a gate term that supervises the measured per-feature gate, and for structure-preserving tasks only, a preservation anchor:
\begin{equation}
\label{eq:total_loss}
\begin{aligned}
\Lcal={}&\underbrace{\lambda_{\mathrm{mmd}}\Lcal_{\mathrm{mmd}}}_{\text{realism}}
+\underbrace{\lambda_{\mathrm{DD\tau}}\Lcal_{\mathrm{DD\tau}}+\lambda_{\mathrm{tv}}\Lcal_{\mathrm{tv}}
+\lambda_{\mathrm{spread}}\Lcal_{\mathrm{spread}}}_{\text{gate}}\\[2pt]
&+\underbrace{\lambda_{\mathrm{struct}}\Lcal_{\mathrm{struct}}}_{\text{preservation}}
+\lambda_{\mathrm{reg}}\Lcal_{\mathrm{reg}}.
\end{aligned}
\end{equation}
Each part is defined formally below, and $P$ denotes the set of latent positions. The supporting
term groups three individually weighted losses,
\[
\lambda_{\mathrm{reg}}\Lcal_{\mathrm{reg}}\equiv\lambda_{\mathrm{dom}}\Lcal_{\mathrm{dom}}
+\lambda_{\mathrm{match}}\Lcal_{\mathrm{match}}^{\mathrm{local}}+\lambda_{\mathrm{type}}\Lcal_{\mathrm{type}}.
\]
For medical translation, $\lambda_{\mathrm{diag}}\Lcal_{\mathrm{diag}}$ is additionally included,
and when the learned noise generator is enabled, $\lambda_{\text{ndist}}\Lcal_{\text{ndist}}$ is
additionally included.

The realism term $\Lcal_{\mathrm{mmd}}$ matches the set of translated outputs to the real
target set with a multi-layer Maximum Mean Discrepancy (MMD), our GAN-free realism driver:
\begin{equation}
\label{eq:mmd_loss}
\begin{aligned}
\Lcal_{\mathrm{mmd}}&=\sum_{\ell}w_\ell\,\mathrm{MMD}^2(\hat B_\ell,B_\ell),\ \text{ with}\\[2pt]
\mathrm{MMD}^2(X,Y)&=\tfrac{1}{n^2}\!\sum_{i,i'}\!k(x_i,x_{i'})
+\tfrac{1}{m^2}\!\sum_{j,j'}\!k(y_j,y_{j'})\\[2pt]
&\quad-\tfrac{2}{nm}\!\sum_{i,j}\!k(x_i,y_j),
\end{aligned}
\end{equation}
with Gaussian kernel $k(a,b){=}\exp(-\lVert a-b\rVert^2/2\sigma^2)$, $\Phi_\ell$ the $\ell$-th Inception
block, $\hat B_\ell{=}\{\Phi_\ell(\hat x_B)\}$ the translated-batch features ($n$ samples),
$B_\ell{=}\{\Phi_\ell(x_B)\}$ the real target features ($m$ samples), and layer weights $w_\ell$.

For the gate, the measured prior supervises the predicted gate through
$\Lcal_{\mathrm{DD\tau}}{=}\lVert\tau-\mathrm{sg}(\tau_{\mathrm{prior}})\rVert^2$, and two light regularizers keep $\tau$ spatially coherent and well-conditioned:
\begin{equation}
\label{eq:tv_spread}
\begin{aligned}
\Lcal_{\mathrm{tv}}&=\tfrac{1}{|P|}\!\sum_{p}\lVert\nabla\tau(p)\rVert_1,\\[2pt]
\Lcal_{\mathrm{spread}}&=\max\!\big(0,\ s^\star-\operatorname{std}_p\tau(p)\big)^2,
\end{aligned}
\end{equation}
with $\nabla$ the spatial gradient and $s^\star$ the target gate spread.

For preservation, used on Type-1 (structure-preserving) tasks only, the $\tau$-gated structure anchor
$\Lcal_{\mathrm{struct}}$ (\Cref{eq:struct}) pins high-$\tau$ positions to the source across a semantic
(DINOv2-feature), a pixel ($L_1$), and an edge granularity, with $\mathrm{sg}(\tau)$ weighting the
anchor toward preserved regions.

Lastly, the three supporting losses are a target-domain classification loss
$\Lcal_{\mathrm{dom}}$, the $k$-nearest-neighbour DINOv2 feature-matching loss $\Lcal_{\mathrm{match}}^{\mathrm{local}}$ of \Cref{eq:matcher_loss},
and a type-aware gate hinge $\Lcal_{\mathrm{type}}$, with weights $\lambda_{\mathrm{dom}}{=}1$,
$\lambda_{\mathrm{match}}{=}1$, and $\lambda_{\mathrm{type}}{=}3$. The type-aware hinge sets the default
gate behavior for Type-1 and Type-2 tasks and is ablated in \Cref{sec:ablation}. Type-1 tasks
also use three auxiliary preservation terms, a latent cycle-consistency loss (translating to the target
and back should recover the source latent), a feature-moment matching loss that aligns the outputs'
deep-feature moments with running target-domain statistics, and a per-frequency-band matching loss that
aligns low- and high-frequency feature statistics with the target. No adversarial or PatchGAN loss is used. The corresponding ablations show that these
auxiliary terms provide only modest gains over the gated-transport core.

\begin{table}[t]
\centering\small
\caption{Role decomposition. Each degree of freedom of the per-sample OT problem
(\Cref{eq:ot_obj}) is handled by a dedicated mechanism.}
\label{tab:roles}
{\renewcommand{\arraystretch}{1.5}\setlength{\tabcolsep}{4pt}%
\begin{tabular}{@{}p{0.29\linewidth}p{0.40\linewidth}p{0.21\linewidth}@{}}
\toprule
Axis & Mechanism & Scope\\
\midrule
Target set $\Mtgt$ & frozen domain-conditional flow $v_\theta$ & global, fixed\\
\addlinespace[2pt]
Per-feature transport cost $\tau$ & learned predictor $\Tau_\psi$ & per-feature\\
\addlinespace[2pt]
$\tau$ spatial pattern & DD$\tau$ prior (distribution distance) & data-driven\\
\addlinespace[2pt]
Preserve/translate regime & content-anchored vs.\ isotropic corruption ($\alpha$) & per-task\\
\addlinespace[2pt]
Trajectory adjustment & constrained correction $C_\phi$ & per-sample\\
\addlinespace[2pt]
Realism (GAN-free) & multi-layer MMD + $k$-NN matcher & distribution\\
\addlinespace[2pt]
Structure preservation & $\tau$-gated structure loss $\Lcal_{\mathrm{struct}}$ & per-feature\\
\addlinespace[2pt]
Hallucination guard & high-$\tau$ source init + diagnosis head & per-feature\\
\bottomrule
\end{tabular}}
\end{table}

\subsection{Optional Extension: A Learned Deterministic Initialization}
\label{sec:lng}
The structure-changing regime is inherently one-to-many, so that with isotropic $\veps\!\sim\!\mathcal N(0,I)$
the gated initialization of \Cref{eq:freq_init} sends one source to a distribution
of valid outputs, one per seed. Results become seed-dependent and invite best-of-$N$ selection. We
remove this ambiguity by learning the source corruption. A small generator $G_\xi$, the Learned Noise
Generator (LNG), emits, per
image, a single deterministic $\veps^\star$, so the translation becomes a reproducible one-to-one map.

To keep $\veps^\star$ a valid corruption rather than an off-manifold shortcut that exploits the
objective, we write it as a residual around the deterministic content-anchored base
$\veps_{\text{struct}}$ (\Cref{eq:content_anchored}) and pin its statistics to the target:
\begin{equation}
\label{eq:lng}
\begin{aligned}
\veps^\star&=\veps_{\text{struct}}+\delta\,G_\xi(\mathbf z_A),\\
\Lcal_{\text{ndist}}&=\big\lVert\mu_c(\veps^\star)-\mu_B\big\rVert^2+\big\lVert\sigma_c(\veps^\star)-\sigma_B\big\rVert^2,
\end{aligned}
\end{equation}
where $\mu_c,\sigma_c$ are per-channel moments and $(\mu_B,\sigma_B)$ the target style-bank moments.
$G_\xi$ is a small zero-initialized convolutional network, so training starts from the known-valid base
($\veps^\star{=}\veps_{\text{struct}}$) and learns only to improve it. It is trained by the
same translation objective (\Cref{sec:objective}). Gradients reach $G_\xi$ through the initialization
path $\mathbf z^0$ and the correction $C_\phi$, while the frozen backbone $v_\theta$ is evaluated with
its velocity detached at every ODE step, so no gradient propagates through it. The reward therefore shapes
$\veps^\star$ toward the corruption that yields the best output for that image, replacing a
best-of-$N$ search into one forward pass. The penalty $\Lcal_{\text{ndist}}$ weighted by
$\lambda_{\text{ndist}}$ constrains $\veps^\star$ to the valid corruption family. We set the residual
scale $\delta{=}1$ and the distribution-validity weight $\lambda_{\text{ndist}}{=}1$. At inference $\veps^\star$ is deterministic, yielding one reproducible output per source with zero seed variance. The main benchmarks use a single random corruption draw for comparability with stochastic baselines, while $G_\xi$
provides a deterministic alternative with comparable realism (\Cref{sec:lng_results}). It modifies only corruption, leaving the frozen flow and gate unchanged.

\section{Experiments}
\label{sec:experiments}
\paragraph{Datasets} We evaluate across five benchmarks spanning both regimes. Three are
structure-preserving (Type~1, only appearance changes): \textbf{TCGA breast
frozen$\to$permanent}~\cite{tcga2013cancer} (histopathology, unpaired same-patient sections),
\textbf{day$\to$night}~\cite{daynight_dataset} (scene relighting), and \textbf{virtual
staining}~\cite{virtualstain_dataset} (unstained$\to$H\&E-stained dermatopathology). Two are
structure-changing (Type~2, geometry
itself changes): \textbf{AFHQ cat$\to$dog}~\cite{choi2020stargan} (natural images, large semantic
gap) and \textbf{face man$\to$woman} (CelebA-HQ~\cite{karras2018progressive}). All images are
processed at $256{\times}256$, and all splits are patient/scene-disjoint and held out. Per-domain
train/test image counts (source/target), used identically for every method, are listed in
\Cref{tab:datasizes}.

\begin{table}[t]
\centering\small
\caption{Per-domain dataset sizes (number of images), from the canonical disjoint splits used for all
methods. ``Source'' is the source domain, ``Target'' the target domain (e.g.\ cat/dog, frozen/permanent).}
\label{tab:datasizes}
\setlength{\tabcolsep}{4pt}
\resizebox{\columnwidth}{!}{%
\begin{tabular}{llcccc}
\toprule
& & \multicolumn{2}{c}{Train} & \multicolumn{2}{c}{Test}\\
\cmidrule(lr){3-4}\cmidrule(lr){5-6}
Dataset & Regime & Source & Target & Source & Target\\
\midrule
AFHQ cat$\to$dog & Type~2 & 5058 & 4669 & 500 & 500\\
CelebA man$\to$woman & Type~2 & 2000 & 2000 & 300 & 300\\
Breast frozen$\to$permanent & Type~1 & 9970 & 9872 & 4475 & 3550\\
day$\to$night & Type~1 & 2000 & 2000 & 300 & 300\\
virtual staining & Type~1 & 6384 & 6420 & 1843 & 3635\\
\bottomrule
\end{tabular}}
\end{table}

\paragraph{Baselines and protocol} We compare against two adversarial translators,
CycleGAN~\cite{zhu2017unpaired} and CUT~\cite{park2020contrastive}, and three diffusion/SDE
translators: an SDEdit-style baseline~\cite{meng2022sdedit} (an SDEdit-style application of our frozen Stage-1 flow rather than a reproduction of the original method), UNSB~\cite{kim2023unpaired}, and EGSDE~\cite{zhao2022egsde}. Every method is retrained on the
identical held-out splits at $256{\times}256$, taken at its best validation-FID checkpoint under a
comparable budget, and scored through one metrics pipeline, so our numbers are directly comparable to
each other but differ from the authors' published values, which used different splits, data volumes, and
FID implementations. Each retrained AFHQ baseline plateaus at its reported validation-FID, confirming convergence. EGSDE is evaluated on the two benchmarks most relevant to its intended application, AFHQ cat$\to$dog and
frozen$\to$permanent translation. Its published AFHQ FID~\cite{zhao2022egsde} was obtained under a different experimental protocol. Under our unified training and evaluation protocol, including retraining and tuning, it achieves a higher FID. Because its sampler requires approximately 10\,s per image, two to three orders of magnitude slower than the other methods, we restrict it to these two benchmarks, one per type.

\paragraph{Metrics} For realism we report the Fr\'echet Inception Distance
(FID)~\cite{heusel2017gans} and the Kernel Inception Distance (KID,
$\times10^3$)~\cite{binkowski2018demystifying}. For structure preservation we report the Nuclei
Preservation Score (NPS), a stain-invariant nuclei-F1 from an H-channel (haematoxylin) detector.
Because an identity map trivially maximizes it (NPS $=1.0$ for a copy), we always read it with the
generated/source nuclei-count ratio, ideal $1.0$, which exposes over- or under-generation. The
evaluation detector is distinct from the StarDist~\cite{schmidt2018stardist} localizer used only for the
$\tau$-override in the controllability review (\Cref{sec:controllability}), so evaluation and control never share a detector. Source-similarity metrics (Learned Perceptual Image Patch Similarity (LPIPS)~\cite{zhang2018unreasonable},
paired DINO cosine, and Multi-Scale Structural Similarity (MS-SSIM)) measure closeness to the input. Because high source-similarity also rewards under-translation,
we use them to characterise the realism-faithfulness trade-off (\Cref{sec:frontier}) rather than as standalone performance metrics.

\paragraph{Implementation details} All images are processed at $256{\times}256$ in the latent space
of a frozen Stable-Diffusion VAE. Stage~1 is a domain-conditional flow-matching U-Net with feature widths of 128, 256, and 512 channels, conditioning dimension 256, including attention at the two coarsest levels. The model is trained for $150$ epochs using Adam optimizer, learning rate of $2{\times}10^{-4}$, and batch size of $8$, after which it is frozen. Stage~2 (the DiT correction
$C_\phi$, the gate predictor $\Tau_\psi$, and the projector) is
trained for $20$ to $90$ epochs depending on the dataset with learning rate of $4{\times}10^{-4}$, and batch size of $8$.
For medical data, the pretrained diagnosis head $h_\psi$ is held frozen and used only to compute $\Lcal_{\mathrm{diag}}$. The feature matcher and DD$\tau$
prior use a frozen DINOv2-S encoder with $q{=}0.95$ quantile normalisation. The seven base weights are
fixed across all datasets: $\lambda_{\mathrm{match}}{=}1.0$, $\lambda_{\mathrm{dom}}{=}1.0$,
$\lambda_{\mathrm{mmd}}{=}35$, $\lambda_{\mathrm{DD\tau}}{=}0.5$, $\lambda_{\mathrm{tv}}{=}0.01$, and
$\lambda_{\mathrm{spread}}{=}3.0$, $\lambda_{\mathrm{type}}{=}3.0$. On the structure-preserving (Type-1)
regime the add-on block is enabled with $\lambda_{\mathrm{struct}}{=}1.0$ (with inner
weights $\lambda_{\mathrm{sem}}{=}1$ and $\lambda_{\mathrm{edge}}{=}1$, and the optional pixel
granularity is disabled, $\lambda_{\mathrm{pix}}{=}0$), $\lambda_{\mathrm{cyc}}{=}0.2$, $\lambda_{\mathrm{fmm}}{=}0.3$,
$\lambda_{\mathrm{band}}{=}1.0$, and the DD$\tau$ weight is raised to
$\lambda_{\mathrm{DD\tau}}{=}1.5$. The medical diagnostic loss weight is $\lambda_{\mathrm{diag}}{=}0.3$. The
$t{=}0$ corruption changes by regime. Type-1 uses the full content anchor ($\alpha{=}1$) with
contrast floor $\tilde\sigma\!\geq\!0.85\,\sigma_c$, whereas Type-2 uses the partial anchor ($\alpha{=}0.5$).
Virtual staining is the exception, using $\alpha{=}0.3$ and $\beta{=}1.0$. All other datasets use $\beta{=}0.5$.
Sampling uses $K{=}16$ Euler steps ($K{=}4$ during training) with classifier-free guidance $w{=}2$ on
natural-image tasks.

\subsection{Natural Images: AFHQ cat$\to$dog}
\label{sec:afhq}
\begin{figure*}[!tb]
\centering
\gridfig{0.90\linewidth}{6}{\small input cat & \small \method{} & \small CycleGAN & \small CUT & \small UNSB & \small SDEdit}{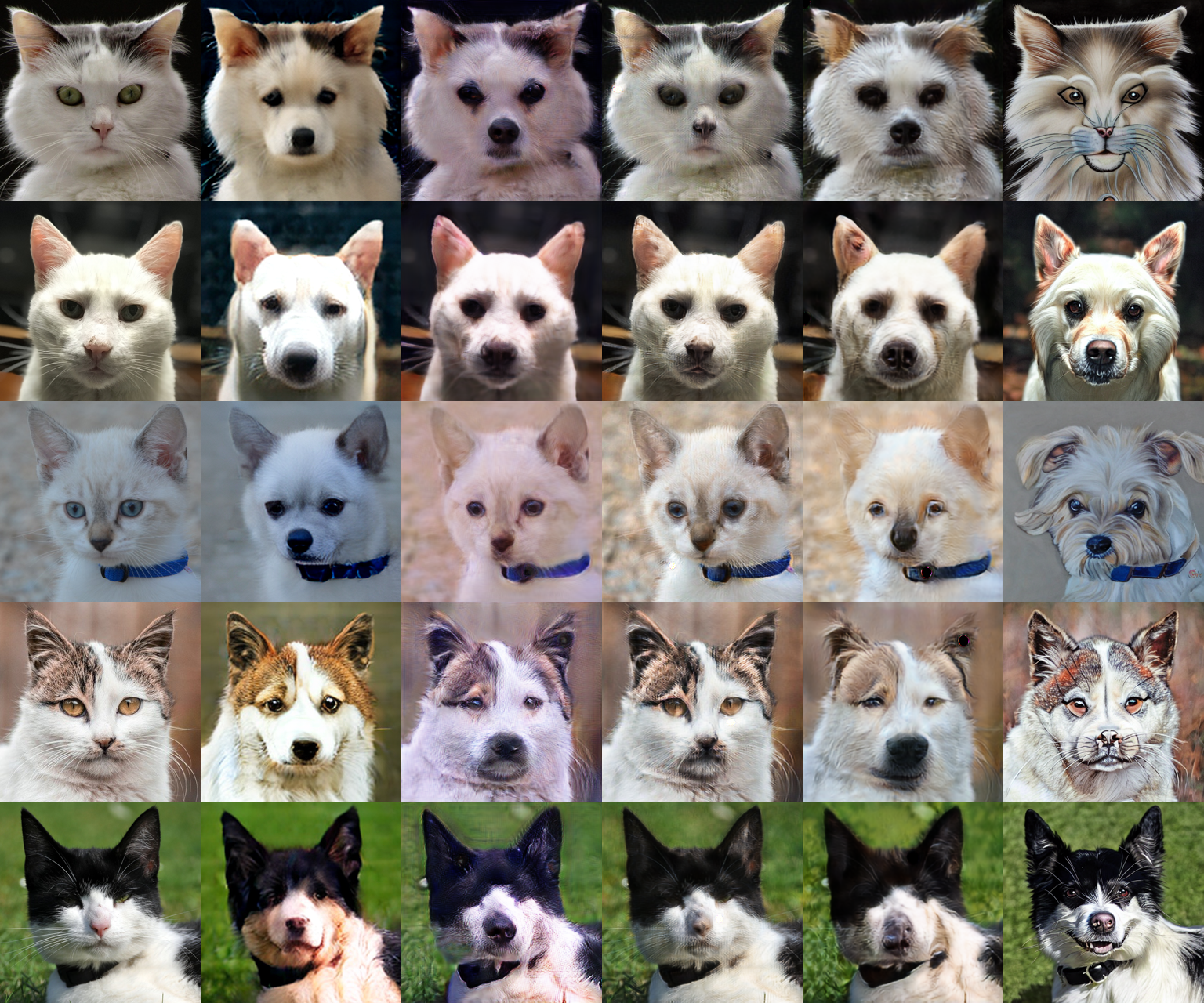}
\caption{Qualitative comparison on AFHQ cat$\to$dog (held-out test cats, one per row). Columns, left
to right: input cat, \method{} (ours), CycleGAN, CUT,
UNSB, SDEdit. \method{} produces realistic
dogs that inherit the source cat's colour, pose and gaze, whereas CUT under-translates (the output stays cat-like), CycleGAN retains cat features, and SDEdit introduces artifacts.}
\label{fig:afhq_compare}
\end{figure*}

\Cref{tab:afhq_results} reports AFHQ cat$\to$dog under the Type-2 configuration. \method{} achieves the lowest FID
(76.9 versus 107.5 for the next best method) and the lowest KID among the evaluated methods. To verify that this improvement is not specific to Inception features, we additionally evaluate the translated images using Inception-independent DINOv2- and CLIP-space Fr\'echet distances, where
\method{} also ranks first. Qualitative examples (\Cref{fig:afhq_compare}) show that \method{} produces realistic dogs while preserving the input cat's coarse appearance and pose, whereas CUT tends to under-translate and the diffusion editors introduce stronger structural changes or artifacts.

\begin{table}[t]
\centering
\caption{AFHQ cat$\to$dog (test cats $\to$ test dogs, clean-FID~\cite{parmar2022aliased}).
$\downarrow$ lower is better.}
\label{tab:afhq_results}
\small
\setlength{\tabcolsep}{4pt}
\begin{tabular}{lcc}
\toprule
Method & FID$\downarrow$ & KID$\times10^3\downarrow$\\
\midrule
CycleGAN~\cite{zhu2017unpaired} & 115.0 & 70.1\\
CUT~\cite{park2020contrastive} & 162.3 & 101.5\\
SDEdit-style Stage-1 flow~\cite{meng2022sdedit} & 107.5 & 65.0\\
UNSB~\cite{kim2023unpaired} & 128.6 & 86.4\\
EGSDE~\cite{zhao2022egsde} & 109.2 & 66.7\\
\method{} (Ours, $\alpha{=}0.5$, $w{=}2$) & \textbf{76.9} & \textbf{27.0}\\
\bottomrule
\end{tabular}
\end{table}

To assess robustness, we retrained \method{} with three independent random seeds (42, 123, and 7), obtaining FID $77.4\pm0.7$, substantially smaller than the margin to competing methods. The anchor strength $\alpha$ further exposes a controllable realism-faithfulness trade-off (\Cref{tab:alpha_sweep}): increasing $\alpha$ improves input faithfulness (paired DINO and LPIPS) while trading off realism (FID). Because Stage~1 is trained with conditioning dropout, classifier-free guidance~\cite{ho2022classifierfree} provides an additional inference-time control. On the unblended flow, increasing the guidance weight to $w{=}3$ improves FID from 89.7 to 78.8, whereas increasing the number of ODE steps beyond the default provides no further benefit. The deployed configuration combines the partial anchor ($\alpha{=}0.5$) with moderate guidance ($w{=}2$), achieving the best overall balance (FID 76.9).

\begin{table}[t]
\centering
\caption{Controlled $\alpha$ sweep on AFHQ (uniform reduced budget, no guidance, only $\alpha$
varies). Increasing the anchor raises input-faithfulness (DINO$\uparrow$, LPIPS$\downarrow$) and trades
realism (FID). Read this table for its monotone trend only. Absolute FID is far higher than the
deployed model because of the reduced budget and disabled guidance. For scale, the same $\alpha{=}0.5$
gives FID $76.9$ at full budget with guidance $w{=}2$ (\Cref{tab:afhq_results}) versus $101.3$ here.}
\label{tab:alpha_sweep}
\small
\setlength{\tabcolsep}{4pt}
\begin{tabular}{ccccc}
\toprule
$\alpha$ & FID$\downarrow$ & KID$\times10^3\downarrow$ & DINO$\uparrow$ & LPIPS$\downarrow$\\
\midrule
0.00 & \textbf{75.3} & 34.6 & 0.189 & 0.520\\
0.25 & 82.3 & \textbf{31.8} & 0.342 & 0.442\\
0.50 & 101.3 & 41.4 & 0.543 & 0.345\\
0.75 & 110.3 & 46.4 & 0.614 & \textbf{0.327}\\
1.00 & 124.2 & 60.6 & \textbf{0.639} & \textbf{0.327}\\
\bottomrule
\end{tabular}
\end{table}

\subsection{Learned Deterministic Initialization}
\label{sec:lng_results}
We evaluate the optional Learned Noise Generator (LNG, \Cref{sec:lng}) on AFHQ cat$\to$dog. LNG replaces the stochastic corruption with a learned deterministic initialization, producing a reproducible one-to-one mapping while matching the realism of stochastic sampling (FID 75.0 versus 75.3). Removing the distribution-validity
penalty $\Lcal_{\mathrm{ndist}}$ degrades FID to 80.0, demonstrating that
constraining $\veps^\star$ to the valid corruption family is essential. LNG therefore removes sampling variability without sacrificing translation quality.

\subsection{Faces: man$\to$woman}
\label{sec:faces}
Our second structure-changing benchmark is CelebA-HQ man$\to$woman. We follow the standard CelebA-HQ binary, appearance-based labels and use ``man$\to$woman'' as shorthand for translating male-labelled to female-labelled facial appearance. These labels are not intended to represent gender identity. The task uses the
same Type-2 configuration as AFHQ, with a partial content anchor ($\alpha{=}0.5$,
\Cref{eq:alpha_blend}) and classifier-free guidance.

Training duration provides an additional control over the realism-identity trade-off. Early in training, the residual correction remains small, and the output is governed primarily by the anchored initialization and frozen flow, producing modest changes while retaining more of the input identity. With continued training, the correction moves the output further toward the female-labelled distribution, improving distributional realism but increasing identity drift. Because the two face domains share substantial structure, prolonged correction can lead to over-translation. We therefore select an early-stopped operating point that balances target-domain appearance with source identity preservation.

\begin{figure*}[!tb]
\centering
\gridfig{0.92\linewidth}{6}{\small input & \small \method{} & \small CycleGAN & \small CUT & \small UNSB & \small SDEdit}{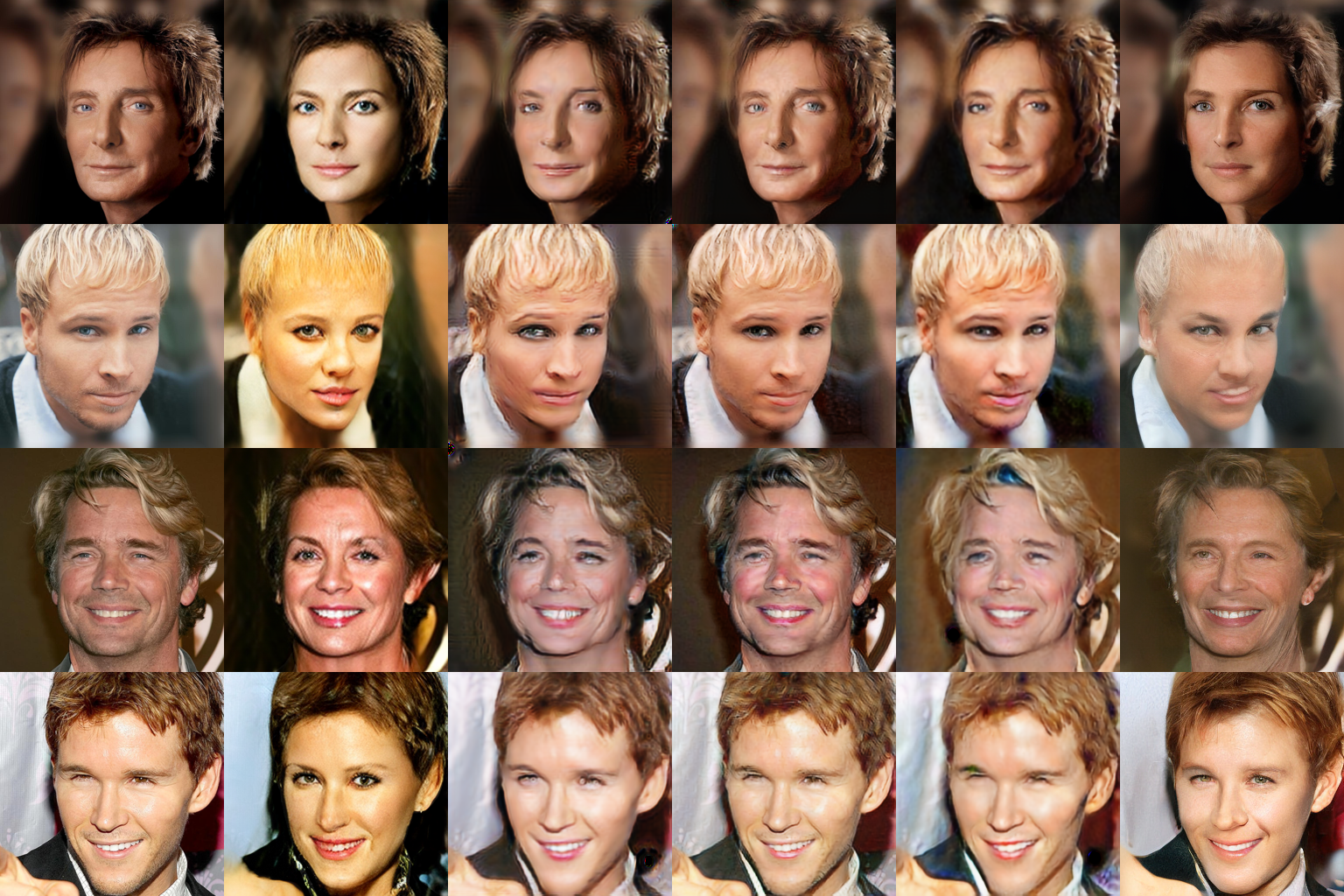}
\caption{Qualitative comparison on CelebA-HQ man$\to$woman (held-out test, one source per row).
Columns, left to right: input, \method{}, CycleGAN, CUT, UNSB,
SDEdit. \method{} alters the target-labelled facial attributes while retaining visually similar pose and
lighting. CUT changes the input little, and the GAN/SDEdit baselines add visible artifacts or shift pose.}
\label{fig:faces_compare}
\end{figure*}

\Cref{tab:faces} shows that \method{} achieves the best quantitative performance among the evaluated methods (FID 90.6, KID $53.9$), improving over the strongest baseline while preserving a more faithful correspondence to the input. \Cref{fig:faces_compare} provides representative qualitative comparisons, illustrating the selected early-stopped operating point.

\begin{table}[t]
\centering
\caption{CelebA-HQ man$\to$woman, $256{\times}256$ held-out, under the shared same-split protocol.}
\label{tab:faces}
\small
\setlength{\tabcolsep}{4pt}
\begin{tabular}{lcc}
\toprule
Method & FID$\downarrow$ & KID$\times10^3\downarrow$\\
\midrule
CycleGAN~\cite{zhu2017unpaired} & 138.4 & 104.0\\
CUT~\cite{park2020contrastive} & 138.8 & 102.8\\
SDEdit-style Stage-1 flow~\cite{meng2022sdedit} & 114.6 & 80.6\\
UNSB~\cite{kim2023unpaired} & 152.6 & 127.5\\
\method{} (Ours) & \textbf{90.6} & \textbf{53.9}\\
\bottomrule
\end{tabular}
\end{table}

\subsection{Histopathology: Breast frozen$\to$permanent}
\label{sec:medical}
\Cref{tab:breast_results} reports breast frozen$\to$permanent translation at $256{\times}256$ on the full held-out test set, with all methods retrained on the identical stratified split. \method{} achieves the lowest Inception FID (51.8) and KID among the evaluated methods, outperforming CUT (56.1), UNSB (65.5), CycleGAN (74.6), SDEdit (139.5), and EGSDE (171.5). It also yields the nuclei-count ratio closest to the ideal value (0.93), indicating the closest agreement between generated and source nuclei counts. This conclusion remains consistent under an independent StarDist detector, for which \method{} obtains a ratio of 1.12 and ranks among the top two methods under both detectors.

The nuclei-count ratio complements FID by quantifying structural preservation and penalizing both missing and hallucinated nuclei. SDEdit degrades nuclear detail through global noising, whereas EGSDE substantially over-generates nuclei, reaching approximately $3.3\times$ the source count. CUT and CycleGAN achieve higher raw nuclei-preservation scores but remain closer to the frozen source, indicating under-translation at the expense of target-domain realism. \method{} therefore provides the most favorable balance between Inception-space realism and near-unity nuclei-count preservation among the evaluated methods. Because under-translating methods may obtain lower distances in pathology-specific or DINOv2/CLIP feature spaces by remaining close to the source, we interpret the histopathology results as a realism-preservation trade-off rather than as an unconditional claim of superior realism.

\begin{table}[t]
\centering
\caption{TCGA breast frozen$\to$permanent at $256{\times}256$, full held-out test. NPS: H-channel
nuclei-F1. Cnt: nuclei-count ratio (generated/source, $1.0$ is ideal, exposes over/under-generation). A high
NPS reached by under-translating (CUT) or a collapse from over-generating nuclei (EGSDE, Cnt $3.3$)
both show in Cnt. Among the compared methods, \method{} best combines realism with a near-ideal nuclei budget. Three-seed
training std for \method{} is $\pm0.16$ FID, far below the margin to the next best method.}
\label{tab:breast_results}
\setlength{\tabcolsep}{4pt}\small
\resizebox{\columnwidth}{!}{%
\begin{tabular}{lcccc}
\toprule
Method & FID$\downarrow$ & KID$\times10^3\downarrow$ & NPS$\uparrow$ & Cnt$\to1$\\
\midrule
CycleGAN~\cite{zhu2017unpaired} & 74.6 & 52.1 & 0.605 & 1.44\\
CUT~\cite{park2020contrastive} & 56.1 & 32.7 & \textbf{0.691} & 1.26\\
SDEdit-style Stage-1 flow~\cite{meng2022sdedit} & 139.5 & 119.8 & 0.380 & 0.82\\
UNSB~\cite{kim2023unpaired} & 65.5 & 42.1 & 0.501 & 2.16\\
EGSDE~\cite{zhao2022egsde} & 171.5 & 162.5 & 0.216 & 3.32\\
\method{} (Ours) & \textbf{51.8} & \textbf{24.7} & 0.582 & \textbf{0.93}\\
\bottomrule
\end{tabular}}
\end{table}

\begin{figure*}[!tb]
\centering
\gridfig{0.98\linewidth}{7}{\small input & \small \method{} & \small CycleGAN & \small CUT & \small UNSB & \small SDEdit & \small EGSDE}{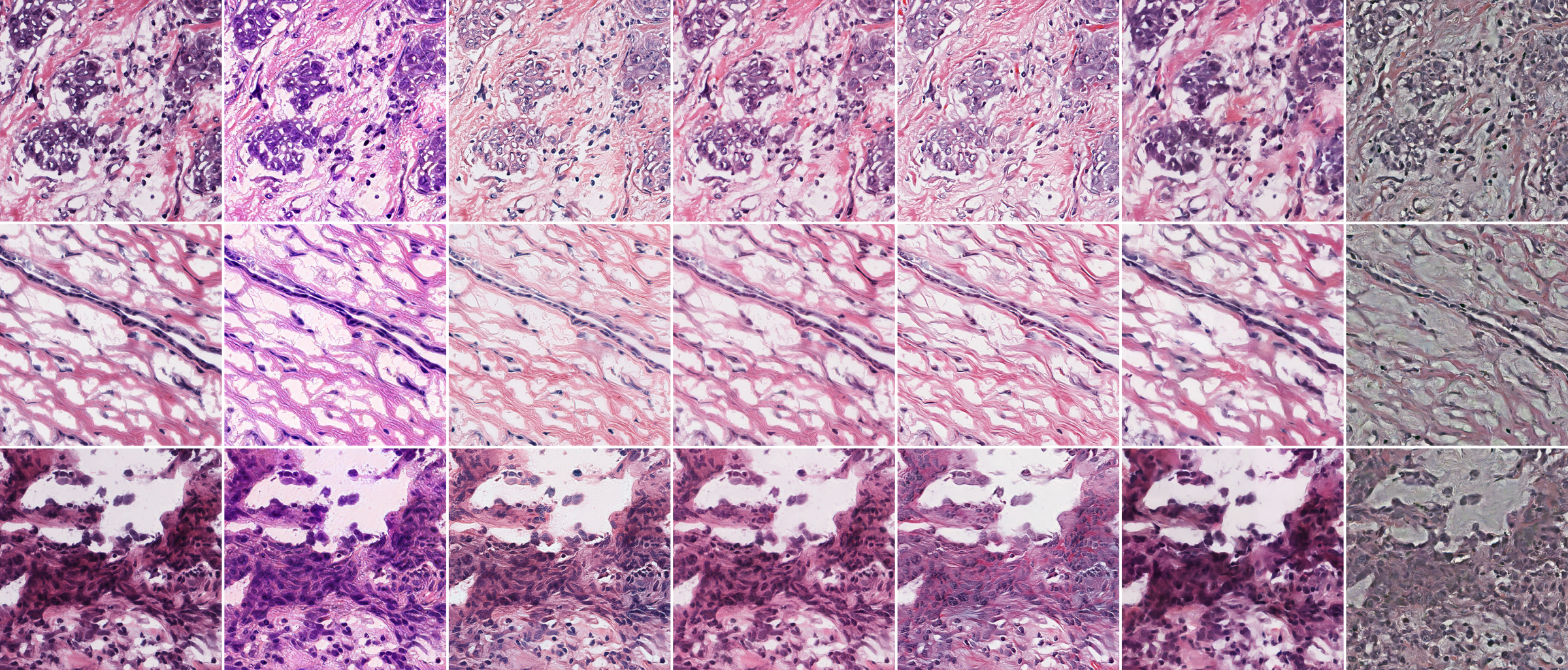}
\caption{Qualitative breast frozen$\to$permanent on reassembled full sections (held-out test,
one section per row, tiles stitched with a shared corruption seed and blended over a 25\% overlap so
no seams appear). Columns, left to right: input frozen section, \method{},
CycleGAN, CUT, UNSB,
SDEdit, EGSDE. \method{} restains toward the permanent
H\&E look while retaining nuclear detail across the field.
EGSDE output is largely desaturated and SDEdit shows degraded nuclei.}
\label{fig:medical_qual}
\end{figure*}

\subsection{Clinical Relevance and Limitations}
\label{sec:clinical}
Representative full-section reconstructions are shown in \Cref{fig:medical_qual}. Tiles are generated using a shared corruption seed and style draw and blended with 25\% overlap, avoiding visible seams while preserving glandular architecture and nuclear organization. These findings are based on automated structural proxies and do not establish clinical utility. A blind pathologist study is therefore required before drawing diagnostic conclusions, and the translated images should be regarded as decision-support outputs rather than replacements for permanent sections.

\subsection{Scene Relighting and Virtual Staining: day$\to$night and unstained$\to$H\&E}
\label{sec:crossdata}
\Cref{tab:crossdata} reports results on two additional structure-preserving benchmarks: day$\to$night relighting and virtual staining from unstained tissue to H\&E. Both use content-anchored initialization, with a full anchor for day$\to$night and a lighter anchor for virtual staining because the unstained source appearance differs substantially from the H\&E target.

\begin{figure*}[!tb]
\centering
\gridfig{0.80\linewidth}{6}{\small input & \small \method{} & \small CycleGAN & \small CUT & \small UNSB & \small SDEdit}{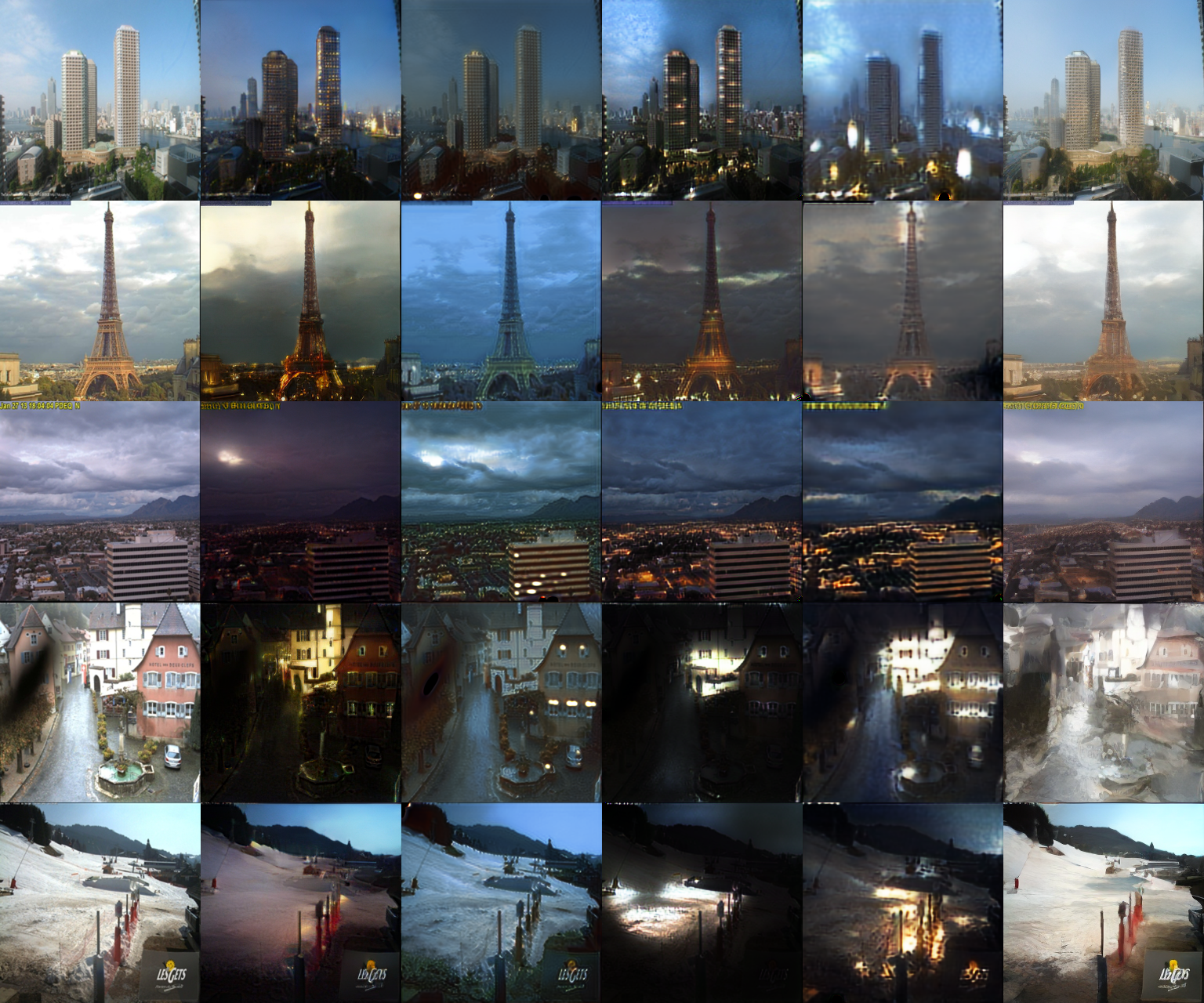}
\caption{Qualitative comparison on day$\to$night. Columns, left
to right: input day image, \method{}, CycleGAN,
CUT, UNSB, SDEdit. \method{}
relights to a convincing night while holding scene geometry largely fixed. UNSB and SDEdit shift structure or
wash out detail.}
\label{fig:dn_compare}
\end{figure*}

On day$\to$night, \method{} achieves the lowest FID (85.9) and KID ($7.6$) among the evaluated methods, improving over CUT, the strongest baseline, at FID 97.7 and KID $14.5$. The qualitative results in \Cref{fig:dn_compare} show that \method{} changes illumination while largely preserving scene geometry, whereas UNSB and SDEdit introduce structural changes or loss of detail.

\begin{figure*}[!tb]
\centering
\gridfig{0.98\linewidth}{6}{\small input & \small ground truth (H\&E) & \small \method{} & \small CUT & \small CycleGAN & \small SDEdit}{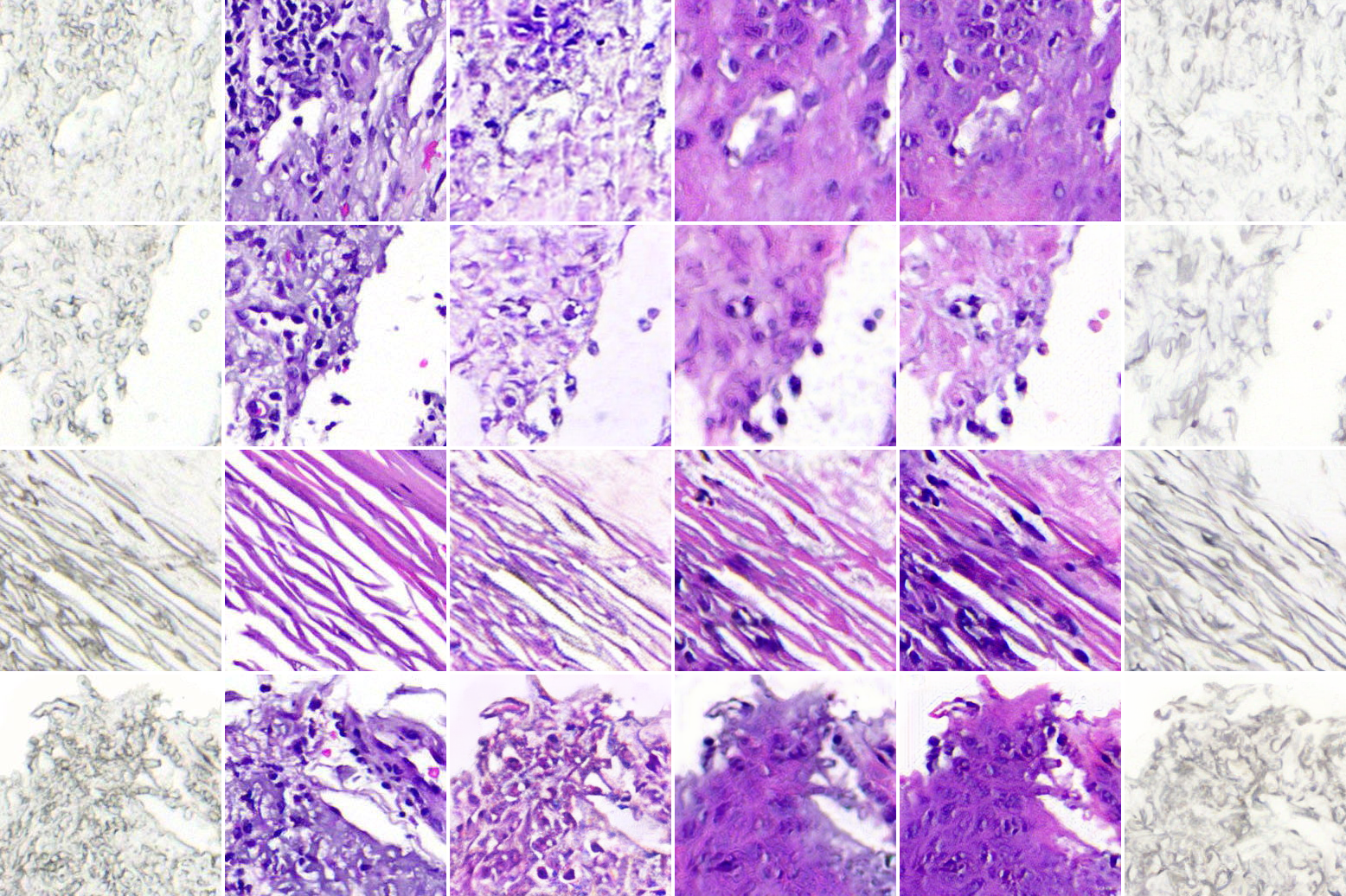}
\caption{Qualitative comparison on virtual staining (unstained$\to$H\&E). Columns: input unstained tissue, the registered H\&E ground truth, \method{}, CUT,
CycleGAN, SDEdit. \method{} introduces an H\&E-like stain while largely preserving the tissue architecture. SDEdit's single global noise level cannot add stain without
destroying tissue, so it barely colors.}
\label{fig:stain_compare}
\end{figure*}

On virtual staining, CycleGAN and \method{} obtain comparable distributional performance: CycleGAN achieves slightly lower FID (49.2 versus 50.8), while \method{} achieves the lower KID ($23.6$ versus $24.9$). We therefore regard the two methods as competitive on this benchmark rather than claiming an overall \method{} advantage. \Cref{fig:stain_compare} shows that \method{} introduces an H\&E-like appearance while preserving tissue architecture. Training remains fully unpaired. The registered H\&E images are included only for qualitative reference and are not used for training, model selection, or hyperparameter tuning.

\begin{table}[t]
\centering
\caption{Type-1 cross-domain results (FID$\downarrow$, KID$\times10^3\downarrow$). \method{} achieves the best score on both metrics on day$\to$night and the best KID on staining, where a fully converged CycleGAN edges FID.}
\label{tab:crossdata}
\setlength{\tabcolsep}{4pt}\small
\resizebox{\columnwidth}{!}{%
\begin{tabular}{lcccc}
\toprule
& \multicolumn{2}{c}{day$\to$night} & \multicolumn{2}{c}{virtual staining}\\
\cmidrule(lr){2-3}\cmidrule(lr){4-5}
Method & FID$\downarrow$ & KID$\times10^3\downarrow$ & FID$\downarrow$ & KID$\times10^3\downarrow$\\
\midrule
CycleGAN~\cite{zhu2017unpaired} & 113.1 & 20.4 & \textbf{49.2} & 24.9\\
CUT~\cite{park2020contrastive} & 97.7 & 14.5 & 71.0 & 48.4\\
SDEdit-style Stage-1 flow~\cite{meng2022sdedit} & 122.8 & 35.3 & 324.4 & 371.5\\
UNSB~\cite{kim2023unpaired} & 131.3 & 40.4 & 100.7 & 78.9\\
\method{} (Ours) & \textbf{85.9} & \textbf{7.6} & 50.8 & \textbf{23.6}\\
\bottomrule
\end{tabular}}
\end{table}

\subsection{Realism-Faithfulness Analysis}
\label{sec:frontier}
Successful unpaired translation requires both target-domain realism and preservation of source information that should remain unchanged. Either criterion alone can be misleading. A method may improve realism by altering excessive source content, whereas high source similarity may result from insufficient translation. We therefore interpret distributional and preservation metrics jointly.

\begin{figure}[t]
\centering
\includegraphics[width=0.92\linewidth]{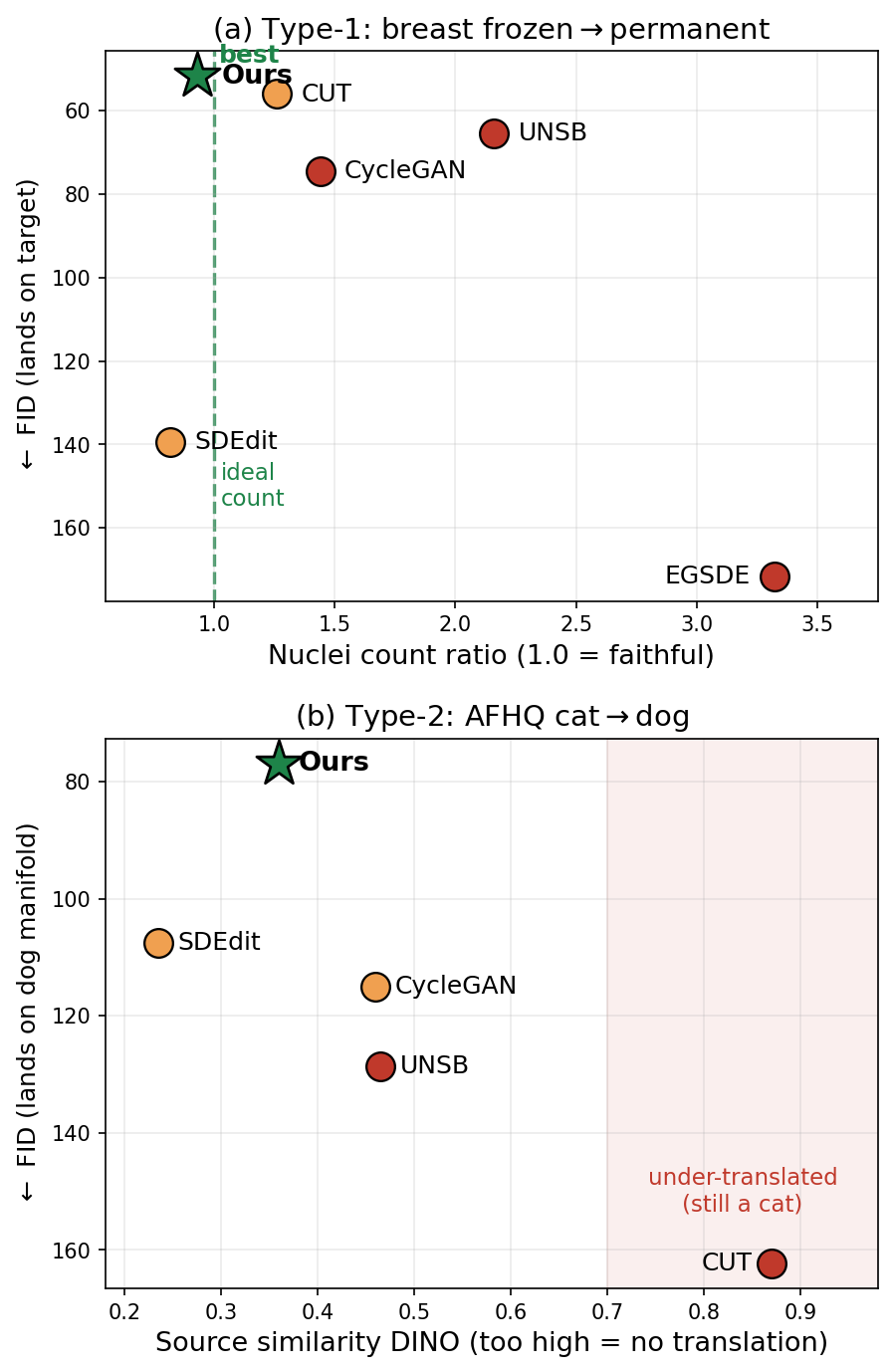}
\caption{Joint realism-faithfulness analysis. The vertical axis reports FID, with lower values indicating greater similarity to the target distribution. (a) For breast frozen$\to$permanent translation, the horizontal axis is the generated-to-source nuclei-count ratio, with 1.0 representing count preservation. (b) For AFHQ cat$\to$dog, the horizontal axis reports source similarity. Unusually high similarity may indicate under-translation because the target requires structural change. \method{} occupies a favorable region of both plots among the evaluated methods.}
\label{fig:frontier}
\end{figure}

\Cref{fig:frontier} presents this analysis for one benchmark from each regime. On breast frozen$\to$permanent translation, \method{} combines the lowest Inception FID with the nuclei-count ratio closest to the ideal value of 1.0. Other methods either remain insufficiently translated or introduce substantial changes in the estimated nuclei count. On AFHQ cat$\to$dog, high source similarity can indicate that the output remains cat-like. Consequently, LPIPS and paired DINO similarity are interpreted together with FID rather than as independent measures of translation quality.

\Cref{tab:similarity} provides the corresponding source-similarity diagnostics. These metrics quantify closeness to the input rather than overall translation quality, and their interpretation therefore depends on the task. High similarity is desirable when geometry should remain fixed, whereas on structure-changing tasks it may instead indicate under-translation. Accordingly, these metrics are interpreted jointly with FID and the realism-preservation analysis of \Cref{fig:frontier}.

\begin{table}[t]
\centering\small
\caption{Source-similarity diagnostics (mean over the held-out test), read jointly with FID. On Type-2
tasks a high value marks under-translation, not a better result, so no value is marked best.}
\label{tab:similarity}
\setlength{\tabcolsep}{4pt}
\resizebox{\columnwidth}{!}{%
\begin{tabular}{llccc}
\toprule
Dataset & Method & LPIPS$\downarrow$ & DINO$\uparrow$ & MS-SSIM$\uparrow$\\
\midrule
\multirow{3}{*}{AFHQ (Type-2)}
 & CUT (under-translates) & 0.16 & 0.87 & 0.92\\
 & CycleGAN & 0.34 & 0.46 & 0.81\\
 & \method{} (Ours) & 0.40 & 0.36 & 0.59\\
\midrule
\multirow{3}{*}{Breast (Type-1)}
 & CUT & 0.09 & 0.95 & 0.94\\
 & CycleGAN & 0.26 & 0.83 & 0.82\\
 & \method{} (Ours) & 0.21 & 0.90 & 0.83\\
\bottomrule
\end{tabular}}
\end{table}

\subsection{Potential for $\tau$-Adaptive Per-Feature Computation}
\label{sec:adaptive_results}
\Cref{tab:nfe} reports the idealized number of active integration steps per feature under the $\tau$-adaptive schedule of \Cref{eq:adaptive_nfe}. Features assigned high preservation values become active later in the trajectory and therefore require fewer integration steps, whereas features requiring substantial translation receive a larger integration budget. The resulting reductions depend on the task and the distribution of predicted gate values, with larger reductions on preservation-dominant breast translation than on structure-changing AFHQ.

These values represent per-feature active-step counts rather than measured wall-clock acceleration. The current dense U-Net and DiT evaluate every spatial position at each integration step, so practical speedups would require structured sparse execution or a backbone capable of skipping inactive regions. \Cref{tab:nfe} should therefore be interpreted as identifying a computational opportunity exposed by the gate, rather than as demonstrating runtime improvement.

\begin{table}[t]
\centering
\caption{Idealized per-feature active-step counts under the $\tau$-adaptive integration schedule. The values do not represent measured wall-clock speedups because the current dense backbone evaluates all spatial locations at every step.}
\label{tab:nfe}
\setlength{\tabcolsep}{4pt}\small
\begin{tabular}{lccc}
\toprule
Dataset & $\bar\tau$ & NFE/region & Saving\\
\midrule
AFHQ (Type~2) & 0.20 & 12.27 & 23.3\%\\
Breast (Type~1) & 0.81 & 2.82 & \textbf{82.4\%}\\
\bottomrule
\end{tabular}
\end{table}

\subsection{Inference-Time, Local Control without Retraining}
\label{sec:controllability}
Because $\tau$ affects sampling only through the initialization and transport gate, the predicted field can be replaced at inference by a gate constructed from an external spatial cue, without retraining the flow or correction network. This enables localized preservation while retaining the default translation behavior elsewhere.

\begin{figure}[!htb]
\centering
\gridfig{0.99\linewidth}{4}{\small input cat & \small \method{} (dog) & \small preserve ``eyes'' & \small preserve ``background''}{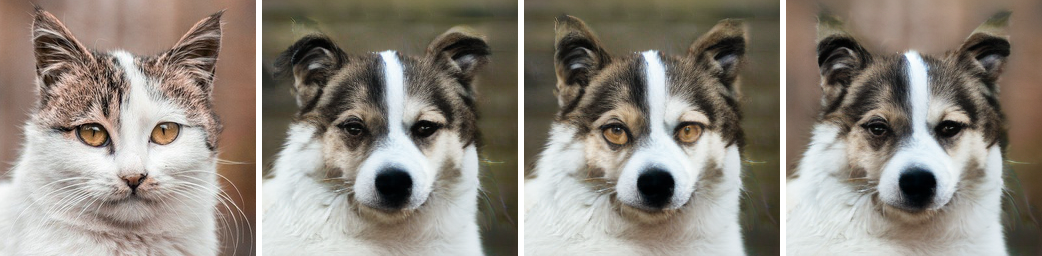}\\[3pt]
\gridfig{0.99\linewidth}{4}{ & & & }{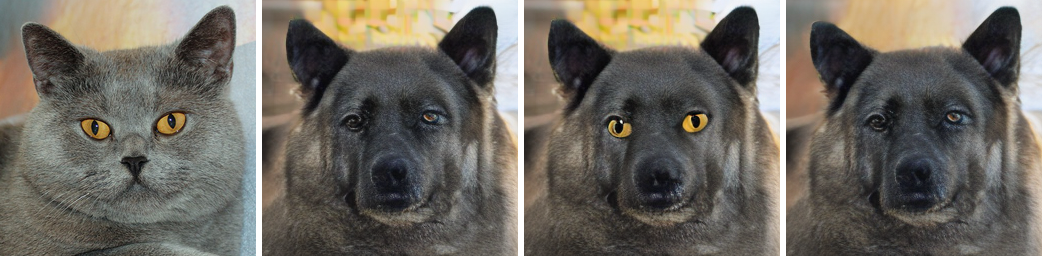}
\caption{Text-guided local preservation during AFHQ cat$\to$dog translation. Each row shows the input, default \method{} output, an ``eyes'' preservation override, and a ``background'' preservation override. The same corruption seed is used across each row so that differences are attributable to the local gate modification.}
\label{fig:control_eyes}
\end{figure}

For natural images, we derive spatial cue from the open-vocabulary segmenter CLIPSeg~\cite{lueddecke2022clipseg}. Given a preservation prompt such as ``eyes'' or ``background,'' its relevance map is thresholded and resampled to the latent resolution. The resulting map increases $\tau$ within the selected region and spatially reduces classifier-free guidance there, while leaving the target-domain guidance unchanged elsewhere. \Cref{fig:control_eyes} shows that ``preserve eyes'' text guidance retains source iris appearance during cat$\to$dog translation, whereas ``preserve background'' retains the surrounding scene while the animal is translated. Across the evaluated cat$\to$dog pairs, the eye-preservation control reduces both eye-region LPIPS and iris-colour error relative to the default translation.

Generic text grounding is unreliable for histopathological structures, so for frozen$\to$permanent translation we instead derive the gate override from a pretrained StarDist nuclei detector~\cite{schmidt2018stardist}. The control mechanism is otherwise unchanged, only the source of the spatial cue differs. Varying the override strength provides a continuous trade-off between source preservation and target-domain realism from a single trained checkpoint.

\subsection{Ablation Study}
\label{sec:ablation}
We evaluate the principal design choices on breast frozen$\to$permanent translation, the benchmark most sensitive to structural preservation. Because the ablation runs use controlled training budgets that differ from the deployed model in \Cref{tab:breast_results}, the results are interpreted comparatively within each experiment rather than against the full-budget benchmark values.

\paragraph{Gate granularity} \Cref{tab:granularity} isolates the effect of gate resolution by comparing global, spatial, and channel-spatial gates while holding the backbone, corruption, correction network, objectives, seed, and training budget fixed. Performance improves as the gate becomes more fine-grained, FID decreases and the nuclei-count ratio approaches its ideal value as control moves from a single global scalar to the full per-feature gate. Although the coarser variants can obtain higher raw NPS, they do so by preserving too much of the source and therefore exhibit poorer target realism and count fidelity. This experiment directly supports the use of channel-spatial rather than global preservation control.

\begin{table}[t]
\centering\small
\caption{Gate-granularity ablation on breast frozen$\to$permanent under a controlled full-data run.
Within this ablation the backbone, corruption, correction, losses, DD$\tau$ prior, seed, training budget,
and evaluation pipeline are held fixed, and only the gate granularity changes. These three models form a
separate controlled experiment, distinct from the main deployed checkpoint reported in
\Cref{tab:breast_results}, so its absolute values are not directly comparable to the deployed $51.8$.
This table should be read for the monotone granularity trend only. Best per column in bold.}
\label{tab:granularity}
\setlength{\tabcolsep}{4pt}
\begin{tabular}{lcccc}
\toprule
Gate & FID$\downarrow$ & KID$\times10^3\downarrow$ & Cnt$\to1$ & NPS$\uparrow$\\
\midrule
Global (scalar $\tau$) & 74.1 & 49.1 & 0.90 & \textbf{0.790}\\
Spatial (per position) & 56.0 & 28.1 & 0.93 & 0.710\\
Per-feature & \textbf{49.1} & \textbf{22.3} & \textbf{0.97} & 0.701\\
\bottomrule
\end{tabular}
\end{table}

\paragraph{Component contributions} \Cref{tab:ablation} evaluates the DD$\tau$ prior, content-anchored corruption, joint gate learning, and norm-constrained correction under a shared reduced budget. DD$\tau$ improves nuclei preservation with little change in FID, indicating that distribution-informed supervision changes the preservation behavior without reducing target realism. Replacing the content-anchored corruption with isotropic noise substantially degrades structural preservation, confirming its importance for Type-1 tasks. Freezing a distilled gate performs worse than joint optimization, while removing the constrained correction produces a different failure mode in which source preservation is retained but target realism deteriorates.

\begin{table}[t]
\centering
\caption{Ablation on breast frozen$\to$permanent (single reduced training budget, compare relative
effects). NPS: H-channel nuclei-F1 (clinical target). Our full configuration is shown in bold, the
only one that performs well on both realism (FID) and preservation (NPS). A single ablation can exceed it
on one axis only by sacrificing the other (for instance dropping the constrained correction raises NPS
to $0.799$ but degrades FID to $128.1$). These values isolate the direction of each effect under a
matched reduced budget and should not be compared numerically with the full-budget deployed checkpoint
of \Cref{tab:breast_results}.}
\label{tab:ablation}
\setlength{\tabcolsep}{4pt}\small
\begin{tabular}{lccc}
\toprule
Configuration & FID$\downarrow$ & NPS$\uparrow$ & Cnt$\to1$\\
\midrule
\method{} full & \textbf{103.6} & 0.750 & \textbf{0.99}\\
w/o DD$\tau$ prior (joint $\tau$ only) & 103.5 & 0.678 & \textbf{0.99}\\
w/o content-anchored corruption & 125.7 & 0.465 & 1.01\\
frozen $\tau$ (distilled, not joint) & 112.3 & 0.628 & 1.10\\
w/o constrained correction ($\beta{\to}\infty$) & 128.1 & \textbf{0.799} & 0.96\\
\bottomrule
\end{tabular}
\end{table}

\paragraph{Core versus auxiliary objectives} \Cref{tab:leancore} compares the full Type-1 objective with a lean configuration that retains the shared base objective and structure anchor but removes the cycle, feature-moment, and frequency-band terms. The lean model remains competitive, while the auxiliary terms provide modest improvements in FID and nuclei-count fidelity. These results indicate that the gated-transport mechanism accounts for most of the performance and that the additional preservation terms provide secondary stabilization on the clinical benchmark.

\begin{table}[t]
\centering\small
\caption{Lean core vs.\ full objective on breast frozen$\to$permanent, at the full (deployed)
training budget. The lean model keeps the seven-term base plus the structure anchor and drops the three
auxiliary Type-1 preservation terms (cycle, feature-moment, per-band). Dropping them costs a small
amount of realism (FID) and preservation, confirming the core, not the auxiliaries, carries the method.
Best per column in bold.}
\label{tab:leancore}
\setlength{\tabcolsep}{4pt}
\resizebox{\columnwidth}{!}{%
\begin{tabular}{lccc}
\toprule
Configuration & FID$\downarrow$ & NPS$\uparrow$ & Cnt$\to1$\\
\midrule
Full \method{} (deployed) & \textbf{51.8} & \textbf{0.582} & \textbf{0.927}\\
Lean core (base $+$ struct anchor, no aux.) & 55.8 & 0.574 & 0.865\\
\bottomrule
\end{tabular}}
\end{table}

\subsection{Consistency Between the Predicted Gate and Feature Transport}
\label{sec:gate_validation}
The ablation studies demonstrate that the predicted gate influences translation quality, but they do not directly examine whether the deployed model transports features in accordance with the learned gate. We therefore analyze the relationship between the predicted gate values and the resulting latent displacement on held-out breast frozen$\to$permanent and AFHQ cat$\to$dog images. For each latent feature, we measure the predicted preservation value $\tau$ and its corresponding source-to-output displacement.

\begin{figure}[t]
\centering
\includegraphics[width=0.96\linewidth]{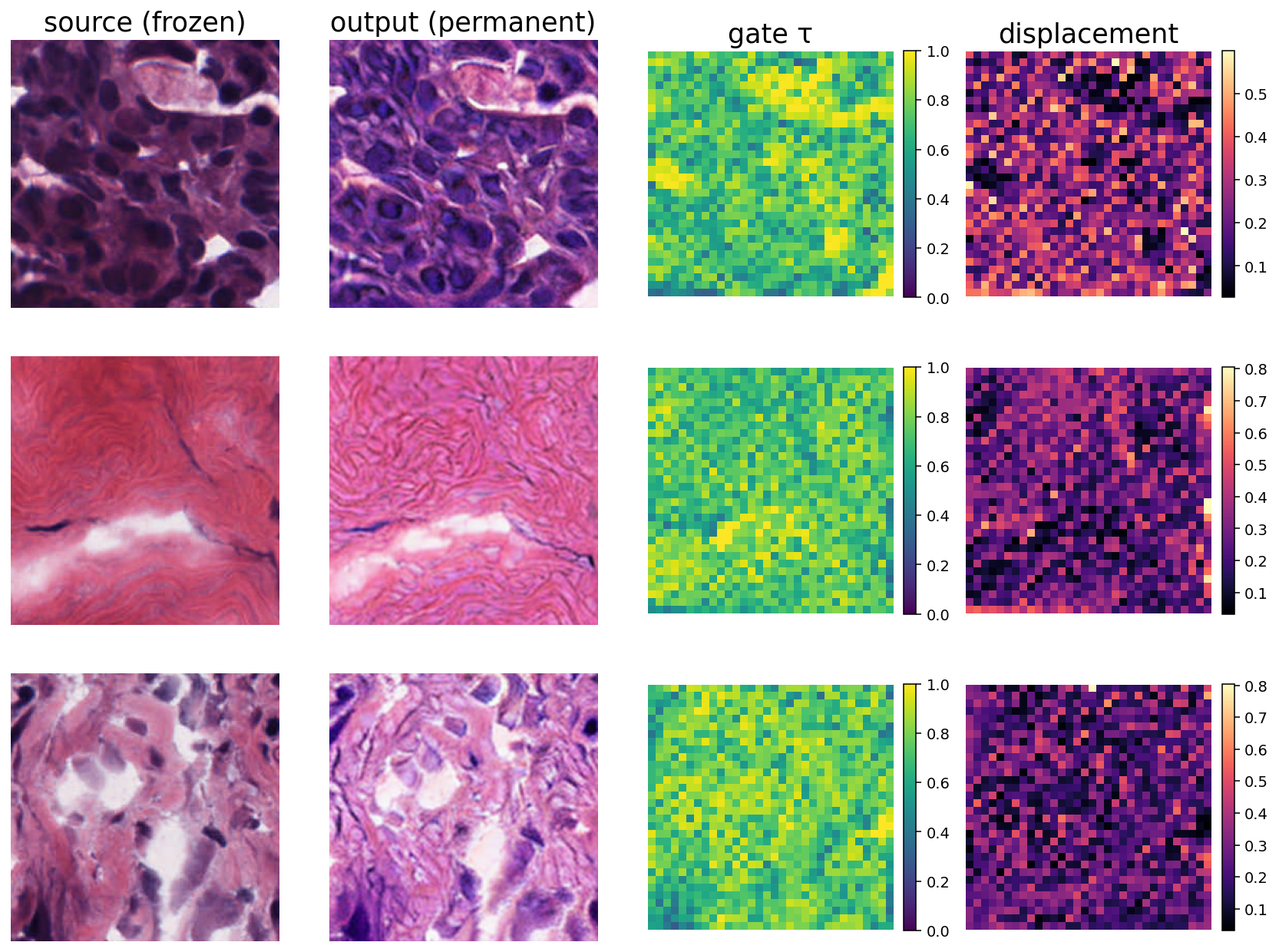}
\caption{Relationship between predicted gate values and latent feature displacement on breast frozen$\to$permanent. Each row shows source, output, the
predicted gate $\tau$, and the actual per-feature displacement $|z_{\text{out}}-z_A|$. In the gate-$\tau$ maps, brighter (yellow) marks higher $\tau$ (more strongly preserved) and darker (purple) lower $\tau$ (more freely translated), on a $[0,1]$ scale. In the displacement maps, brighter marks larger per-feature displacement, so high-$\tau$ regions coincide with low displacement.}
\label{fig:gate_validation}
\end{figure}

\Cref{fig:gate_validation} shows that features assigned larger gate values generally undergo smaller latent displacements, whereas features with smaller gate values receive larger updates. This trend is strongest for breast frozen$\to$permanent translation, where much of the image is intended to remain unchanged, and is weaker for AFHQ, where larger structural changes are required across the image.

Although this analysis does not establish semantic correctness of the predicted gate, it provides empirical evidence that the learned gate is meaningfully associated with the amount of transport performed by the deployed model. Together with the gate-granularity, DD$\tau$, and inference-time override experiments, these results support the interpretation of the gate as a feature-dependent preservation signal rather than merely an auxiliary prediction.

\section{Discussion}
\label{sec:discussion}
\method{} couples a measured per-feature preservation gate with task-matched initialization. As a result, feature transport varies systematically with the predicted gate on held-out data (\Cref{sec:gate_validation}).

\paragraph{Realism and faithfulness together} On benchmarks where realism is the primary objective, \method{} achieves the lowest FID (AFHQ 76.9, breast 51.8). For histopathology, where preserving tissue structure is equally important, we additionally report nuclei preservation, the nuclei-count ratio, and the realism-faithfulness frontier (\Cref{sec:frontier}). Together, these results show that \method{} achieves a favorable realism-preservation balance while allowing task-specific operating points at inference.
\paragraph{Limitations} The study has four main limitations. First, the gate is validated through ablation, displacement, and inference-time override analyses rather than against ground-truth spatial change masks. Second, the gated dynamics are an empirical design and do not inherit a formal distributional guarantee from the frozen flow. Similarly, the $\tau$-adaptive compute analysis reports idealized reductions in active integration rather than measured wall-clock acceleration. Third, the pathology evaluation relies on automated structural proxies and does not replace blinded pathologist assessment or clinical validation. Finally, the evaluation is limited by finite-sample metrics, single-run estimates on some benchmarks, separate training for each domain pair, and the $256{\times}256$ latent-resolution setting. Broader validation at higher resolutions, across additional domains, and with more extensive controllability studies remains future work.
\paragraph{Future work} \Cref{sec:adaptive_results} demonstrates that the $\tau$-adaptive solver concentrates integration where transport is needed while reducing computation in preserved regions using a fixed threshold on the predicted gate. A natural extension is to replace this heuristic with a learned, differentiable controller that allocates integration steps on a per-feature basis, jointly optimizing translation quality and computational cost. Features far from the target distribution would receive more integration steps, whereas preserved features would receive few or none. Building on prior work in input-adaptive computation and spatially sparse processing~\cite{yoshai2025sparse}, such a formulation could learn an input-dependent compute allocation while leaving the frozen flow and correction network unchanged.
\section{Conclusion}
\label{sec:conclusion}
We formulate unpaired image translation as selective per-feature transport guided by a measured preservation prior (DD$\tau$), implemented over a frozen flow with a norm-constrained, GAN-free correction. Because preserved features remain anchored to the real source and undergo little transport, a single trained model provides a controllable realism-preservation trade-off through inference-time text or detector cues while allocating transport primarily to regions requiring change. Across five benchmarks spanning both structure-preserving and structure-changing tasks, evaluated under a unified protocol without pretrained text-to-image generators, \method{} achieves the lowest Inception FID and KID on four benchmarks and competitive performance on the fifth. On histopathology, it also yields the nuclei-count ratio closest to the ideal value, indicating improved structural preservation while maintaining realistic translation.

\section*{CRediT authorship contribution statement}
\textbf{Elad Yoshai:} Conceptualization, Methodology, Software, Validation, Formal analysis,
Investigation, Writing - original draft, Visualization.
\textbf{Natan T. Shaked:} Writing - original draft, Resources, Funding acquisition, Supervision.

\section*{Declaration of competing interest}
The authors declare that they have no known competing financial interests or personal relationships
that could have appeared to influence the work reported in this paper.

\section*{Funding}
This research received no specific grant from any funding agency in the public, commercial, or
not-for-profit sectors.

\section*{Ethics}
This study used only publicly available, de-identified data: TCGA histopathology (GDC portal, under
the TCGA data-use terms), the public AFHQ, CelebA-HQ and day$\to$night benchmarks, and a public
dermatopathology archive. No new human or animal data were collected, and no institutional review
board approval was required.

\section*{Data availability}
The TCGA histopathology data are publicly available from the GDC portal. AFHQ, CelebA-HQ and
day$\to$night are public benchmarks. The virtual-staining slides are from a public
dermatopathology archive~\cite{virtualstain_dataset}. The training source code, trained weights, split manifests, generated test outputs, and self-contained evaluation code can be released upon publication.

\section*{Statistical protocol}
All methods were evaluated under the shared protocol of \Cref{sec:experiments}, so the reported results are directly comparable point estimates. To characterize the sampling variability of FID, we bootstrap-resampled the held-out test sets 1000 times and recomputed FID for each resample. The resulting standard deviations were small (0.4 for breast, 1.3 for AFHQ on \method{}, and 1.5 for AFHQ on CUT), indicating that the observed inter-method differences substantially exceed the estimated sampling variability on these benchmarks. Training variability was assessed using three independent training seeds on AFHQ and breast. The remaining benchmarks are reported as single-run point estimates.

\bibliography{references}

\end{document}